\documentclass[preprint, numbers, sort&compress]{elsarticle}

\usepackage{a4wide}
\usepackage{graphicx}
\usepackage{amsmath,amssymb,amsfonts}
\usepackage{amsthm}
\usepackage{titlecaps}
\usepackage{booktabs}
\usepackage{float}

\usepackage{xcolor}%
\usepackage[hidelinks]{hyperref}

\usepackage{tikz}

\usepackage{siunitx}
\usepackage{microtype}
\usepackage{arydshln}
\usepackage[capitalize,noabbrev]{cleveref}

\newcommand{\mat}[1]{\boldsymbol{#1}}
\renewcommand{\vec}[1]{\mathbf{#1}}

\newcommand{\NORMAL}[2]{$\mathcal{N}\left(#1,\,#2\right)$}
\newcommand{\PRIOR}[1]{$p\left(#1\right)$}

\begin{document}

\begin{frontmatter}
    \journal{Mechanical Systems and Signal Processing}

    \title{A Latent Restoring Force Approach to Nonlinear System Identification}

    \author{T.J.\ Rogers$^1$}
    \ead{tim.rogers@sheffield.ac.uk}
    \author{T.\ Friis$^2$}

    \address{$^1$Dynamics Research Group, Department of Mechanical Engineering, University of Sheffield, Mappin Street, Sheffield, S1 3JD,UK}
    \address{$^2$Department of Civil Engineering, Technical University of Denmark, Brovej, 2800 Kgs., Lyngby, Denmark}

    \begin{abstract}
        Identification of nonlinear dynamic systems remains a significant challenge across engineering. This work suggests an approach based on Bayesian filtering to extract and identify the contribution of an unknown nonlinear term in the system which can be seen as an alternative viewpoint on restoring force surface type approaches. To achieve this identification, the contribution which is the nonlinear restoring force is modelled, initially, as a Gaussian process in time. That Gaussian process is converted into a state-space model and combined with the linear dynamic component of the system. Then, by inference of the filtering and smoothing distributions, the internal states of the system and the nonlinear restoring force can be extracted. In possession of these states a nonlinear model can be constructed. The approach is demonstrated to be effective in both a simulated case study and on an experimental benchmark dataset.
    \end{abstract}
    
    \begin{keyword}
        Bayesian, Nonlinear System Identification, Gaussian Process, Grey-box Model
    \end{keyword}

\end{frontmatter}


\section{Introduction}
\label{sec:introduction}

Identification of nonlinear dynamic systems remains an outstanding challenge across engineering. %
Although this paper will focus on mechanical applications, nonlinear systems emerge across a range of disciplines from electrical and control systems to bioengineering. %
In identifying these systems there are generally two aims for the engineer; first, to gain insight into the physical phenomena driving the behaviour of that system, and secondly to make predictions of the response of that system to new inputs. %
Owing to its broad applicability and challenge, nonlinear system identification has seen a great deal of research interest sustained over many years, for a good review see \cite{noel2017nonlinear}. %
Although often presented, in a mechanical setting, as second-order nonlinear differential equations, a nonlinear dynamic system can be considered equivalently in a state-space framework. %
The distinction being that several internal states are modelled explicitly, as is a defined ``measurement model'' which relates these states to some observed quantity; this model also allows for the concepts of both process noise and measurement noise. %
Mathematically, the general continuous time state-space model is shown in \cref{eq:ssm}.
\begin{subequations}
    \begin{align}
        \dot{\vec{x}}(t) &= f(\vec{x}(t),\vec{u}(t),\vec{v}(t))\\
        \vec{y}(t) &= g(\vec{x}(t),\vec{u}(t),\vec{w}(t))
    \end{align}
    \label{eq:ssm}
\end{subequations}
\noindent where $\vec{x}(t)$ is a vector of internal states at time $t$ whose time derivatives are contained in $\dot{\vec{x}}(t)$. %
These derivatives are given by some function $f(\cdot)$ of the states $\vec{x}(t)$, some known external inputs $\vec{u}(t)$ and some process noise $\vec{v}(t)$. %
The internal states may not be directly observed, instead a quantity $\vec{y}(t)$ is the measured signal at time $t$ which itself is a function $g(\cdot)$ of the states $\vec{x}(t)$, some known external inputs $\vec{u}(t)$ and some measurement noise $\vec{w}(t)$.


The work shown in this paper will consider one specific task which a practitioner may be interested in when confronted with a nonlinear system. %
That is to identify from measured inputs and outputs a nonlinear system where the governing equation of motion is unknown \emph{a priori}. %
In particular, a modern viewpoint on a classic nonparametric approach will be shown, that is a Bayesian treatment of the nonlinear restoring force surface method of \citet{masri1979rfs} when limited measurement information is available. %
The methodology presented is of particular value in cases where measurement noise means that numerical estimates of the internal states of the system (displacement, velocity, etc.) are poor.

The key concept employed in this work is to hypothesise that the unknown \emph{nonlinear restoring force} can be modelled as a \emph{Gaussian process} (GP) in time (\cite{o1978curve,rasmussen2006gaussian}) and therefore incorporated into a Gaussian process \emph{latent force model} (LFM) (\cite{alvarez2009latent,hartikainen2012sequential}). %
In other words the contribution of the unknown nonlinear terms to the equation of motion is replaced with a GP as a function in time which recovers the unknown nonlinear restoring force. %
The GP provides a prior over a function \cite{rasmussen2006gaussian}, in this case the signal in time which is the nonlinear restoring force, the function is updated in a Bayesian manner as observations of the system response are made to recover a posterior distribution over the potential contribution of the nonlinear restoring force. %
The recovered nonlinear restoring force along with estimates of the displacement and velocity can then be used to infer the unknown terms in the nonlinear equation of motion in isolation. %
Therefore, the contribution of this paper is to show how the Gaussian process Latent Force Model (GPLFM) might be employed in a novel manner to aid system identification of nonlinear systems with nonlinear terms which are unknown \emph{a priori}, through separating their contribution from that of the linear system components.

\subsection{Related Work}
\label{sec:related_work}

The scope of nonlinear system identification is far reaching and cross-disciplinary. %
Within a mechanical context, which is the focus of this work, good reviews of progress being made can be found in \cite{kerschen2007nonlinear} and more recently, in the subsequent work of \cite{noel2017nonlinear}. %
The range of applied technologies has been broad ranging from physically derived models to machine learning approaches. %
Examples of physically grounded methods include the control-based continuation approach of \cite{barton2017control} or Approximate Bayesian Computation to learn models and parameters in \cite{abdessalem2018model} or applications of nonlinear filtering algorithms \cite{chatzi2009unscented}. %
From a purely data-driven perspective, as early as \cite{masri1993identification} and \cite{chance1996higher} black-box tools such as neural networks were being considered as models of nonlinear dynamic systems and this continues with the current popularity of deep learning approaches; e.g.\ \cite{ljung2020deep}.
\cite{masri1993identification} in fact builds on the restoring force (\cite{masri1979rfs}) approach by using a neural network to learn a model of the nonlinear function rather than the Chebyshev polynomials used in the original work. %
The general approach seen in \cite{masri1979rfs} led to widespread use, for example \cite{al1990application,mohammad1992direct}, and in \cite{kerschen2006past} is cited as one of the major families of approach in time-domain identification. %
The power of the restoring force approach in \cite{masri1979rfs} is evidenced by its inclusion in the updated review paper of \cite{noel2017nonlinear} which includes numerous examples of its ongoing application. %
The machine learning community has more generally considered a range of problems within the scope of nonlinear dynamic system identification. 
For example, methodologies such as the neural network approaches of \cite{chen2018neural} and \cite{raissi2018hidden} or GPs in the case of \cite{damianou2013deep}.

Of particular relevance to the work shown in this paper is the work of \cite{alvarez2009latent} with the introduction of the GPLFM. %
This is a mechanistically inspired machine learning model as discussed in the previous section. %
In \cite{alvarez2013linear} applications in gene expression and the diffusion of pollutants are discussed as the dynamic processes being considered. %
Within a mechanical engineering context the ideas of the GPLFM have been exploited for solving the joint input-state-parameter problem by the authors in \cite{rogers2018oma} and \cite{rogers2020application}. %
\cite{nayek2019gaussian} also discussed the use of such a model for input estimation with known system parameters.
Here the task considered was to infer, from a set of measured accelerations, the unmeasured states (displacements and velocities), parameters (mass, stiffness and damping matrices) and inputs (excitation forces) in a joint framework based on the GPLFM. %
The approach was extended in \cite{rogers2020bayesian} to an input-state estimation method for nonlinear systems where the parameters and form of the system were known but the inputs were not.

\usetikzlibrary{calc}
\usetikzlibrary{arrows.meta}
{
    \renewcommand{\familydefault}{\sfdefault}

    \begin{figure}

    \begin{tikzpicture}

    \tikzstyle{flowchart} = [align=center, minimum size=3cm, draw, line width=0.05cm, rounded corners=0.2cm ]
    \tikzstyle{flowarrow} = [-{Triangle[round, scale=0.75]}, line width=0.1cm]

    \node[flowchart] (a) at (0,0) {\textbf{A}: Assume model as\\[0.2cm] $m\ddot{z} + f(z,\dot{z})=U(t)$};
    \node[flowchart] (d) at ($(a) + (0,4)$) {Collect data\\[0.2cm] \includegraphics[width=4cm]{figures/duffing/acceleration_measurement}};
    \node[flowchart] (b) at ($(a) + (0,-4)$) {\textbf{B}: Further assume\\[0.2cm] $m\ddot{z} + c\dot{z} + kz + \hat{f}(z,\dot{z})=U(t)$};
    \node[flowchart] (c) at ($(b) + (0,-4)$) {\textbf{C}: Consider\\[0.2cm] $\hat{f}(z,\dot{z}) \sim \mathcal{GP}\left(0, k\left(t,t^\prime\right)\right)$};
    \node[flowchart] (e) at ($(c) + (8,0)$) {\textbf{D}: Convert dynamics and\\ GP to linear SSM\\[0.2cm] $\begin{aligned}
        \begin{bmatrix}
            \dot{\vec{x}} \\ \dot{\vec{f}}
        \end{bmatrix} = \begin{bmatrix}
            \mat{A_{c,s}} & \mat{B_{c,f}}\\
            \vec{0} & \mat{A_{c,f}}
        \end{bmatrix}\begin{bmatrix}
            \vec{x} \\ \vec{f}
        \end{bmatrix} + \begin{bmatrix}
            \mat{B_{c,s}}\\\vec{0}
        \end{bmatrix}U(t) + L\nu(t)
    \end{aligned}$};
    \node[flowchart] (f) at ($(e) + (0,4)$) {\textbf{E}: Perform MCMC over the SSM\\ Computation $\mathcal{O}\left(KT\right)$};
    \node[flowchart] (g) at ($(f) + (-3,4)$) {\textbf{F.1}: Recover parameters\\ \includegraphics[width=4cm]{figures/duffing/param_posteriors}};
    \node[flowchart] (h) at ($(f) + (2,4)$) {\textbf{F.2}: Recover states\\ \includegraphics[width=4cm]{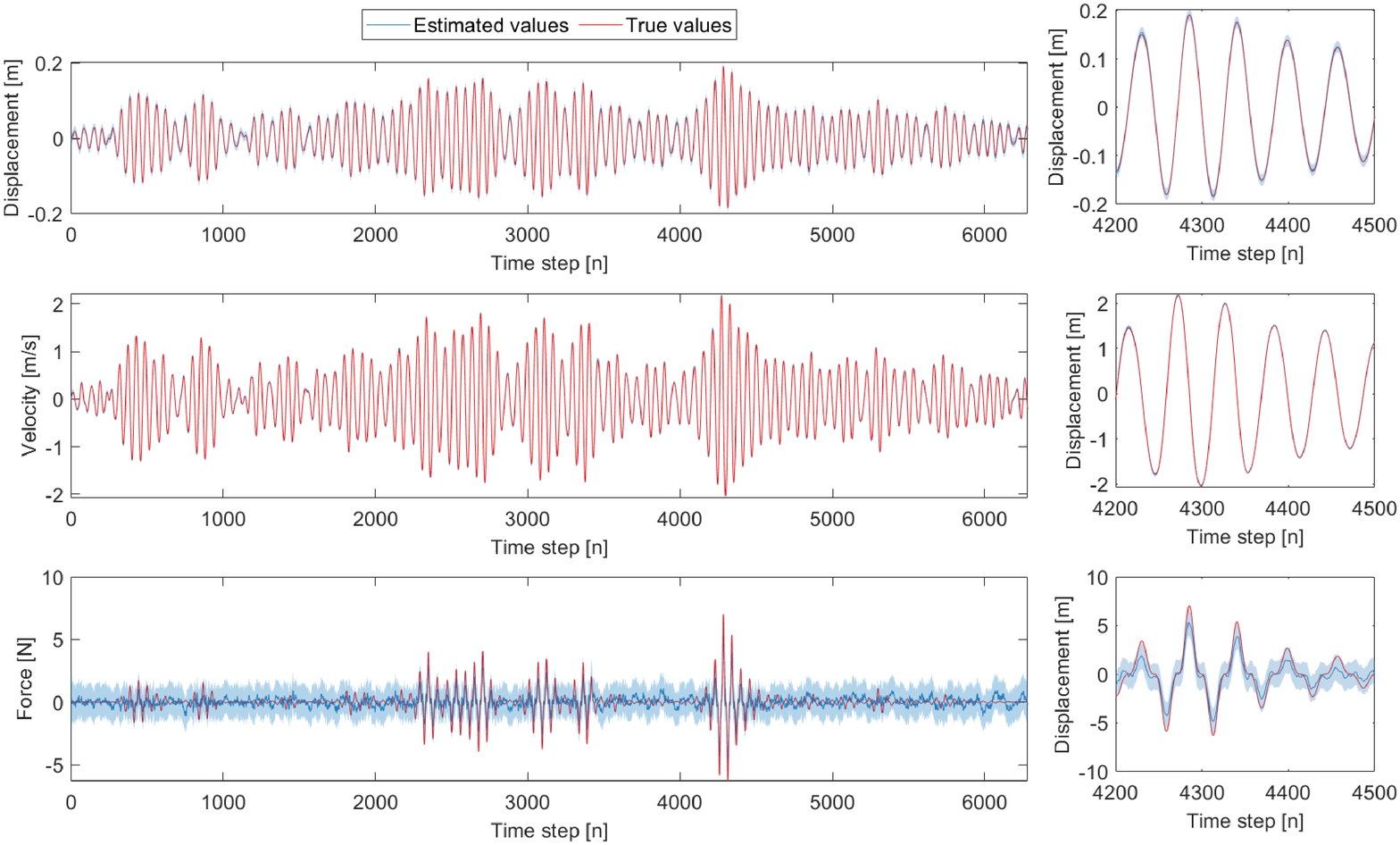}};
    \node[flowchart] (i) at ($(h) + (0,4)$) {\textbf{G}: Extract Nonlinearity\\ \includegraphics[width=4cm]{figures/duffing/nonlinear_force_data}};
    \node[flowchart] (j) at ($(g) + (0,4)$) {\textbf{H}: Fit nonlinearity with\\ \emph{static} model\\ \includegraphics[width=4cm]{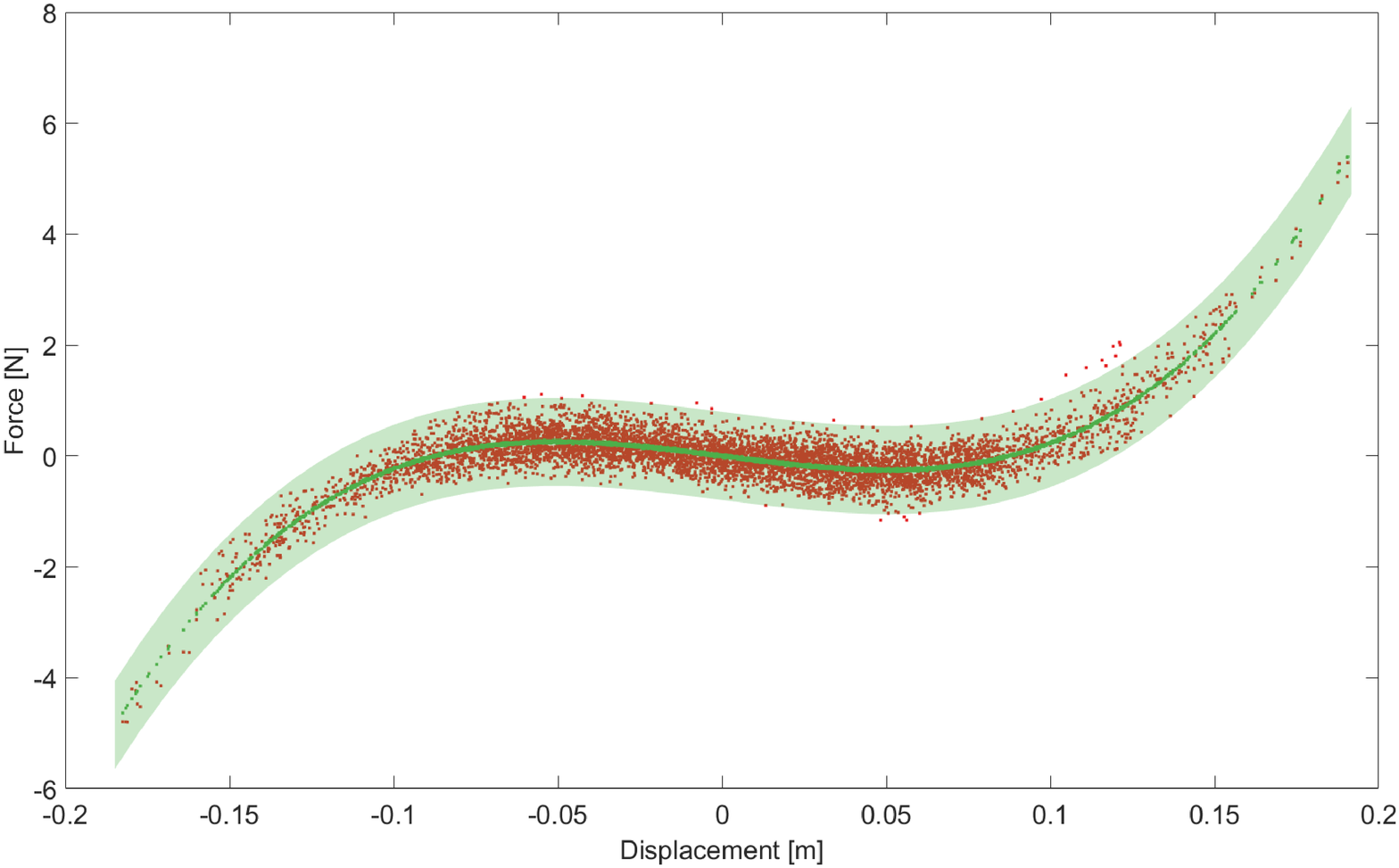}};

    \draw[flowarrow] (d) -- (a);
    \draw[flowarrow] (a) -- (b);
    \draw[flowarrow] (b) -- (c);
    \draw[flowarrow] (c) -- (e);
    \draw[flowarrow] (e) -- (f);
    \draw[flowarrow] (f) -- (g);
    \draw[flowarrow] (f) -- (h);
    \draw[flowarrow] (h) -- (i);
    \draw[flowarrow] (i) -- (j);

    \end{tikzpicture}

    \caption{Process for GPLFM based restoring force identification.}
    \label{fig:flowchart}

    \end{figure}

} 

\subsection{Overview of Proposed Approach}
An overview of the proposed methodology presented in this work is shown in \cref{fig:flowchart}. %
The key steps of the method and main contributions of this work are those shown in blocks C, D and E. %
Specifically, the restoring force estimation problem is recast as a state-estimation problem where the unknown component (in time) is modelled as a GP. %
This gives rise to a linear SSM over which inference can be performed, since the SSM remains linear computation time is greatly reduced. %
In addition, the identification aims to decouple the dynamic model from the fitting of the nonlinear term (blocks G and H). %
By inferring the states and the missing contribution to the system along with the parameters (F.1 and F.2) it is possible to extract the \emph{static} form of the nonlinearity, i.e.\ it is possible to plot the restoring force curve/surface as in block G. %
Fitting a static model to this data is far less demanding that fitting and comparing multiple nonlinear dynamic models. %
Having given this very general overview of the proposed approach, the specific construction of the model can now be presented.

\section{From Restoring Force Surfaces to Latent Restoring Forces}
\label{sec:theory}

Before discussing the particular approach that is adopted in this work, it is worth reviewing the classic restoring force surface approach of \citet{masri1979rfs}. %
The ongoing applicability of this method is a testament to both its intuitive approach and the quality of that original work. %
The authors in that paper note that for a single-degree-of-freedom (SDOF) system the differential equation of motion can be written as,
\begin{equation}
    m\ddot{z} + f(z,\dot{z}) = U(t)
    \label{eq:nonlinear_system}
\end{equation}
\noindent for an object of mass $m$ with displacement $z$, velocity\footnote{Throughout this paper the notation $\dot{(\cdot)}$ is adopted to indicate the first derivate with respect to time and likewise $\ddot{(\cdot)}$ for the second derivative.} $\dot{z}$ and acceleration $\ddot{z}$; forced by a known signal $U(t)$. %
Then the function $f(z,\dot{z})$ is termed the \emph{restoring force} of the system. %
In the special case where $f(z,\dot{z}) = c\dot{z} + kz$, for scalar coefficients $c$ and $k$, the equation of motion of a linear SDOF mass-spring-damper system is recovered. %
However, the case of interest is when this restoring force is generated by some unknown, possibly nonlinear, function. %
\citet{masri1979rfs} observe that if the mass $m$, acceleration $\ddot{z}$ and excitation $U(t)$ are known, for example from an experiment study, the equation of motion can be rearranged to find this unknown function $f(z,\dot{z})$.
\begin{equation}
    f(z,\dot{z}) = U(t) - m\ddot{z}
\end{equation}
Having made this rearrangement, a functional form for the restoring force is estimated. %
While there are many possible options for this modelling, the one highlighted in the original paper is the use of Chebyshev orthogonal polynomials. %
Through fitting this function $f(z,\dot{z})$, given the observations of the excitation $U(t)$ and the inertial term $m\ddot{z}$, it is possible to then predict the response of the nonlinear system to some new excitation signal or to inspect the form of this function for physical insight.

In practice, this methodology can perform very well, however, one drawback that may be encountered is the need to measure or estimate the displacement, velocity and acceleration of the system in order to fit the function $f(z,\dot{z})$. %
This may increase the cost of testing to introduce more sensor modalities or require the user to resort to some estimation method for the missing quantities. %
By far the most common measurement, of a mechanical dynamic system, is the acceleration of the mass $\ddot{z}$. %
From this, the use of numerical integration techniques could be employed to estimate the displacement and velocity of the mass. %
The issue with this approach is that error is often introduced during this procedure, which may distort results or require further post-processing. %
This difficulty increases significantly with noise on the measurements which are collected.
\citet{worden1990data} discusses in further detail some of the challenges associated with employing these numerical methods within the restoring force surface methodology. 

Inspired by the restoring force surface approach, this paper looks to overcome the issues associated with obtaining measurements or estimates of the displacement, velocity and acceleration by estimating them within a Bayesian state-space framework. %
The aim then being to recover simultaneously, the parameters of the underlying linear model, the unmeasured quantities and the missing nonlinear restoring force.

\subsection{Gaussian Process Latent Force Models}\label{sec:GPLFM}

\citet{alvarez2009latent} introduced a model where a data-driven regression model, a Gaussian process, is linked to a mechanistic representation of the data --- an ordinary differential equation (ODE). %
For example to model data generated by a second-order ODE in the form,
\begin{equation}
    m\ddot{z} + c\dot{z} + kz = U(t), \qquad U(t)\sim\mathcal{GP}(0,k(t,t^\prime))
    \label{eq:GPLFM}
\end{equation}
\noindent where the unknown forcing on the differential equation $U(t)$ is modelled as a GP with zero mean and a covariance function $k(t,t^\prime)$ which is stationary and depends on time.

The original purpose of the GPLFM was to increase the flexibility and expressive power of the GP in modelling data. %
This is shown to work well on a number of examples. %
One limiting factor to the approach shown in \citet{alvarez2009latent} is the computational burden of computing the latent force model, this is in some way addressed in \cite{alvarez2009sparse} through the introduction of sparse inference over the GP\footnote{There are a number of approaches for sparse computationally efficient inference in GPs, one of the most common frameworks is that which employs a number of pseudo-points or inducing points, a good overview is provided in \cite{bui2016unifying} which also links two of the most popular methods.}. %
An alternative approach to increasing the computational efficiency of the latent force model is made available through work of \cite{hartikainen2010kalman}, who show that a GP in time with zero mean and a stationary covariance function can be rewritten as a stochastic differential equation (SDE) which has an equivalent form as a linear Gaussian state-space model (LGSSM). %
Once in possession of this LGSSM, inference can be performed using the well known Kalman filter \cite{kalman1960new} and Rauch-Tung-Streibel (RTS) smoother \cite{rauch1965maximum}. %
Importantly, \cite{hartikainen2010kalman} show that the smoothing solution to the obtained model is equivalent to the posterior solution of the full GP under the usual Gaussian likelihood. %
Once able to model the forcing of the system as a GP in time, which can be written as linear state-space model, it is a sensible step to link this with a state-space representation of the dynamic system on the left-hand side of \cref{eq:GPLFM} and perform inference over the resulting joint state-space model (SSM).
This combination of the two ideas was shown in \cite{hartikainen2012sequential} which allows $\mathcal{O}(T)$ inference of the latent force model to take place via the Kalman filtering, RTS smoothing algorithms. %
For the full details of these approaches, the reader is referred to the original papers; \cite{alvarez2009latent} as an introduction followed by \cite{hartikainen2010kalman} then \cite{hartikainen2012sequential} for the implementation as an LGSSM. %

\subsection{A Latent Restoring Force Model}\label{sec:GPLRF}
The approach of the GPLFM assumes a linear ODE driven by a GP which is unmeasured and must be inferred. %
This lends itself naturally to problems in mechanical engineering such as the joint input-state-parameter estimation problem where it has been previously applied, see \cref{sec:related_work}. %
However, in this work a different problem is considered. %
It is assumed that measurements of the input to the system $U(t)$ are available with the exact form of the ODE describing the physical behaviour of the system being the unknown.

It remains to show that a similar approach to that of the GPLFM can be adopted to infer the unmeasured states and the nonlinear restoring force of the system simultaneously in a state-space framework. %
Recalling \cref{eq:nonlinear_system}, the model being considered is one of a (potentially nonlinear) second order differential equation with an unknown restoring force function $f(z,\dot{z})$. %
Assuming that there is some linear component, this can be rewritten as,
\begin{equation}
    m\ddot{z} + c\dot{z} + kz + \hat{f}(z,\dot{z}) = U(t)
    \label{eq:eom_rflfm}
\end{equation}
\noindent where the nonlinear component of the restoring force is represented by a new but still unknown function $\hat{f}(z,\dot{z})$. %
It will also be considered that only measurement of the acceleration is available from testing, although an equivalent method could be written down trivially for different sensor modalities, e.g. velocity measurements from a laser Doppler vibrometer.
It is necessary to infer the unknown displacement $z$, velocity $\dot{z}$ and the unknown restoring force function $f(z,\dot{z}) = kz + c\dot{z} + \hat{f}(z,\dot{z})$ simultaneously as all three quantities are unmeasured.

Inference over the coupled quantities (states, parameters and the unknown nonlinear function) directly is difficult.
For example, it is possible to consider multiple possible forms for $f(z,\dot{z})$ and calibrate each of those models, eventually selecting the one which is optimal in some sense. %
This approach has been seen in the literature, an example already discussed is \citet{abdessalem2018model} using ABC to perform joint model and parameter selection. %
While effective, it --- along with many other direct approaches --- requires multiple nonlinear simulations which impose a high computational cost. %

Therefore, a strategy is proposed in the same spirit as the GPLFM where it is assumed that $\hat{f}(z,\dot{z})$ may be represented as a GP in time with zero mean and a stationary kernel,
\begin{equation}
    m\ddot{z} + c\dot{z} + kz + \hat{f}(z,\dot{z}) = U(t), \qquad \hat{f}(z,\dot{z})\sim\mathcal{GP}(0,k(t,t^\prime))
    \label{eq:GPLFM_RF}
\end{equation}
\noindent This construction of the model gives rise to two related benefits; firstly inference may now be performed in a joint manner over $z$, $\dot{z}$ and $\hat{f}(z,\dot{z})$; secondly, the nonlinear system is approximated by a linear SSM leading to significant computational advantages. %
These computational gains mean that Bayesian inference over the parameters of the model: the mass, stiffness and damping; as well as the hyperparameters of the GP is feasible. %
Having conducted Bayesian inference over the linear dynamic model which is far less computationally intensive, the nonlinear (and linear if desired) components of the model can be fit in a \emph{static} manner. %
In other words, extracting $z$, $\dot{z}$ and $\hat{f}(z,\dot{z})$ allows the ODE describing the system to be learnt without the need for further costly nonlinear simulations of the full dynamic model.

The formulation of such a model can now be shown; although the finer details of representing the GP in a state-space form will not be covered. %
These details can be found in the original papers of \cite{hartikainen2010kalman,hartikainen2012sequential} or in the context of structural dynamics in \cite{rogers2020application}. %
The process of inferring the model begins by converting the system into a state space form in continuous time given a state vector $\vec{x} = \begin{bmatrix}
    z & \dot{z}
\end{bmatrix}^\text{T}$, as in \cref{eq:cont_system_ssm}.
\begin{equation}
    \begin{aligned}
        \dot{\vec{x}} &= \underbrace{\begin{bmatrix}
            0 & 1 \\ -\frac{k}{m} & -\frac{c}{m}
        \end{bmatrix}}_{\mat{A_{c,s}}}\vec{x} + \underbrace{\begin{bmatrix}
            0 \\ \frac{1}{m}
        \end{bmatrix}}_{\mat{B_{c,s}}}\left(U(t) - \hat{f}(z,\dot{z})\right) \\
        \vec{y} &= \mat{C}\vec{x} + \mat{D}(U(t) - \hat{f}(z,\dot{z}))
    \end{aligned}
    \label{eq:cont_system_ssm}
\end{equation}

It can be observed that this has the form of a standard continuous-time state-space model, with the dynamics governed by $\mat{A_{c,s}}$ and $\mat{B_{c,s}}$, and the observations made via $\mat{C}$ and $\mat{D}$. %
These matrices $\mat{C}$ and $\mat{D}$ will be dependent on the particular sensing modality used; for example, if observing acceleration $\mat{C} = \begin{bmatrix}-k/m & -c/m\end{bmatrix}$ and $\mat{D} = 1/m$.

Since the nonlinear restoring force $\hat{f}(z,\dot{z})$ is not a known or measured quantity, it must be modelled approximately using some suitable prior assumptions which can be updated through the observation data. %
As discussed, this is achieved through modelling $\hat{f}(z,\dot{z})$ as a Gaussian process in time with a zero mean and a stationary kernel $k(t,t^\prime)$. 
Notationally, the model of the nonlinear restoring force is comprised of a state vector $\vec{f}$, a state transition matrix $\mat{A_{c,f}}$ and a white noise process $\nu(t)$ with a spectral density $q$ entering on one of the states of $\vec{f}$: $$ \dot{\vec{f}} = \mat{A_{c,f}}\vec{f} + \mat{L}\nu(t)$$ %
The size of $\vec{f}$, structure of the $\mat{A_{c,f}}$ matrix and spectral density $q$ are fully defined by the form of the kernel $k(t,t^\prime)$ via its spectral density $S(\omega)$ \cite{hartikainen2010kalman}. %
For certain classes of kernel, including the widely used Mat\'ern type \cite{stein2012interpolation}, these quantities are available in a closed form which gives rise to a finite dimensional linear state-space model. %
The derivation of the matrix $\mat{A_{c,f}}$ from a particular kernel is not included in this work for brevity, it can be found in full in the literature, for example see \cite{hartikainen2010kalman,hartikainen2012sequential,nayek2019gaussian,rogers2020application,sarkka2019applied}.

What remains is the combination of these two models into a single SSM, such that inference can be performed jointly over the unknown states of the system (the displacement and velocity) and the unknown nonlinear restoring force. %
This involves one modification to \cref{eq:cont_system_ssm}, that being to move the nonlinear restoring force $\hat{f}(z,\dot{z})$ into the state vector $\vec{x}$, creating a new extended state-transition matrix $\mat{A_c}$ and control matrix $\mat{B_c}$. %
The observation matrices $\mat{C}$ and $\mat{D}$ will also be updated to account for the relation of the new extended state vector to the measured quantity or quantities. %
Appending the state vector $\vec{x}$ with the additional states related to the GP representation of the nonlinear restoring force gives,
\begin{equation}
    \begin{aligned}
        \begin{bmatrix}
            \dot{\vec{x}} \\ \dot{\vec{f}}
        \end{bmatrix} = \begin{bmatrix}
            \mat{A_{c,s}} & \mat{B_{c,f}}\\
            \vec{0} & \mat{A_{c,f}}
        \end{bmatrix}\begin{bmatrix}
            \vec{x} \\ \vec{f}
        \end{bmatrix} + \begin{bmatrix}
            \mat{B_{c,s}}\\\vec{0}
        \end{bmatrix}U(t) + \mat{L}\nu(t)
    \end{aligned}
    \label{eq:nonlinear_restoring_force_lfm}
\end{equation}

\noindent where $\mat{B_{c,f}}$ models how the force will act on the physical dynamic system. %
In the SDOF cases considered in this work, this is a matrix of zeros except in the lower left-hand corner where it equals $1/m$, this can be seen from the rearrangement of \cref{eq:eom_rflfm}. %
The matrix $\mat{L}$ controls how the white noise process $\nu(t)$ enters the system, in this case it is a column vector where all elements are zero except the final one which is unity. %
The formulation in \cref{eq:nonlinear_restoring_force_lfm} provides a linear system approximation of the nonlinear system, where the nonlinear restoring force is represented by a Gaussian process in time. %
The object of interest is the smoothing distribution of the LGSSM which can be efficiently computed via the Kalman filter and RTS smoother, this distribution is the posterior over the internal states of the model which include the ``physical states'' and the additional states related to the GP which represents the nonlinear restoring force. %
In order to practically employ these methods, the system must first be discretised which is a standard procedure and can be found in many good textbooks, for example see \cite{ljung1998system,sarkka2013bayesian}. %

Once the posterior distributions over the states have been recovered, these may be used to build a model of the nonlinear component of the system in a manner similar to \cite{masri1979rfs}. %
Samples of the states can be drawn which capture the uncertainty in the estimated displacement and velocities as well as in the unknown nonlinear restoring force. %
In possession of these samples, the task is then to fit a model of the nonlinearity, this step is not covered in detail in this work since many options exist. %
For instance, one could adopt a polynomial method (or orthogonal polynomial) as in the original work of \cite{masri1979rfs}. %
Alternatively, a more physically-informed approach could be taken by inspection of the system itself or the nonlinear restoring force now isolated from the rest of the dynamics. %
If not in possession of strong prior knowledge, then a purely ``black-box'' approach may be taken, for example a neural network or GP could be used to model the nonlinear function. %
After a model is developed of the nonlinearity, on the basis of the state estimates, this can be used in future prediction tasks. %

At this point it is worth noting that the authors would recommend the use of numerically stable square-root forms of the Kalman filter and RTS smoother. %
This can overcome issues when the measurement noise in the system is very low. %
Specifically, the formulations shown in \cite{wills2012estimation} were used in the case studies shown in this paper, including the procedure for sampling from the posterior distributions over the states. %
Without the use of these more stable methods, it was found that the variance in the samples of the states could be artificially inflated or in extreme cases lead to over/underflow issues in the computation.

\section{Case Studies}
\label{sec:case_studies}

This section will present two examples of how the proposed methodology could be applied to the identification of a nonlinear system. %
The first, a simulated study with a Duffing oscillator; second, an experimental dataset known as the Silverbox produced by \cite{wigren2013three} which electrically exhibits response similar to a Duffing oscillator. %
This will demonstrate the effectiveness of the method in recovering a model in the case where it can be compared to a ground truth and also when attempting to identify a physical test object which is more representative of an industrial system identification task.

\subsection{Simulated Duffing System}
\label{sec:sim_duff}

The Duffing oscillator is arguably one of the simplest interesting nonlinear systems that an engineer may wish to study. %
Its equation of motion is identical to that of a linear mass-sping-damper system except the addition of a term which is cubic in the displacement, i.e.,

\begin{equation}
    m\ddot{z} + c\dot{z} + kz + k_3 z^3 = U(t)
    \label{eq:duffing}
\end{equation}

A system following \cref{eq:duffing} is simulated using a Newmark integration scheme \cite{newmark1959method} with a mass $m=1\si{\kilo\gram}$, linear stiffness $k=100\si{\newton\per\metre}$, viscous damping $c=0.4\si{\newton\second\per\metre}$ and cubic stiffness $k_3=1000\si{\newton\per\metre\cubed}$. %
The system is simulated for 12566 time steps with a sample frequency of 100\si{\hertz}. %
The excitation used is a random phase multisine based on 1000 sampled frequencies and amplitudes from a JONSWAP wave spectrum \cite{ochi2005ocean} with $H_s=2.5$ and $T_p=1$.

\begin{figure}[h]
    \centering
    \includegraphics[width=0.8\textwidth]{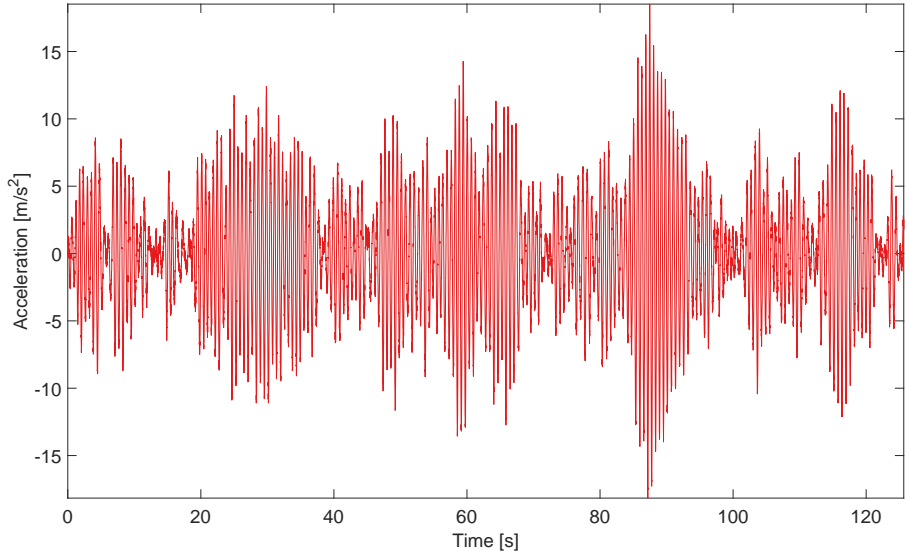}
    \caption{Acceleration measurement from the simulated Duffing oscillator}
    \label{fig:duff_acc_sim}
\end{figure}

It is assumed that only measurements of the acceleration of the oscillator are available and these are polluted with an artificial noise through addition of i.i.d. samples from a Gaussian distribution with zero mean and a standard deviation close to 5\% of the standard deviation of the noise free signal, this gives a noise variance $R=0.05$. %
This noise addition replicates the sort of behaviour that may be expected when testing a physical system. %
\cref{fig:duff_acc_sim} shows the time domain signal which is used as the ``measurement'' in this simulated Duffing study. %

It can now be shown how the framework proposed will allow identification of this nonlinear dynamic system. %
From \cref{eq:GPLFM_RF} and \cref{eq:duffing} it can be seen that this system is well suited to identification in the manner described in this paper. %
Therefore, it is trivial to form \cref{eq:nonlinear_restoring_force_lfm} for this SDOF nonlinear system based on the known state-space form of a linear SDOF system (for example see \cite{ljung1998system}, chapter 4) which gives the matrices $\mat{A_{c,s}}$ and $\mat{B_{c,f}}$. %
What remains is to specify a form for the GP which is chosen to model the nonlinear restoring force in the system. %
In this case a Mat\'ern kernel is chosen with a roughness of $1/2$. %
In possession of this kernel, it is possible to write down the form of $\mat{A_{c,f}}$ and the spectral density of the white noise process $\nu(t)$, as functions of the two kernel hyperparameters: the signal variance $\sigma_f^2$ and length scale $\ell$, see \cite{sarkka2019applied},

\begin{figure}[h]
    \centering
    \includegraphics[width=0.8\textwidth]{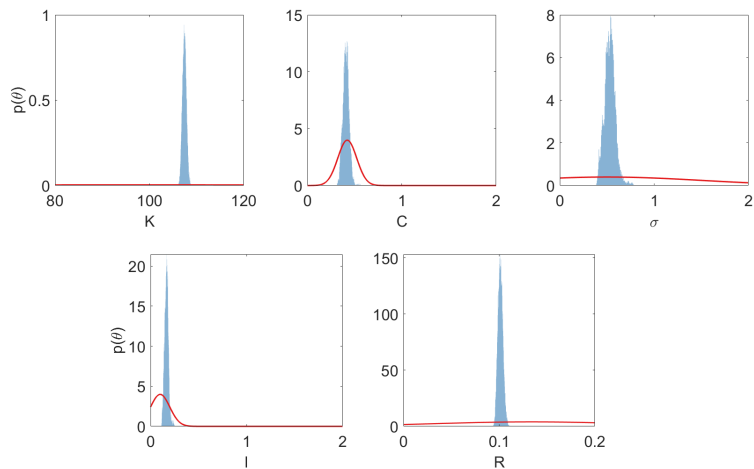}
    \caption{Parameter posteriors from the MCMC inference over the state-space model. Solid (red) lines indicate the probability density function of the prior and the histograms (blue) are built from samples of the posterior.}
    \label{fig:duff_posteriors}
\end{figure}

Given this construction of the model there are a number of unknown quantities which must be inferred. %
It is assumed that the mass of the oscillator is known but the remaining system parameters, the damping coefficient $c$ and linear stiffness $k$ are not. %
The nonlinear stiffness $k_3$ is not modelled at this stage but is also assumed unknown. %
There are two hyperparameters of the GP which are not known, as discussed there are $\sigma_f^2$ and $\ell$. %
Finally, it is also assumed that the measurement noise variance $R$ is unknown. %
This leaves five free (hyper)parameters which must be identified. %
The posterior distributions of these unknowns are estimated by means of a Markov Chain Monte Carlo (MCMC) inference where the (unnormalised log) likelihood is available in closed form from the Kalman filter and the priors are defined independently as follows in \cref{tab:C1_priors} after perturbing the true values, noting that the ground truth values for the hyperparameters of the GP are not available.

\begin{table}[h]
    \centering
    \caption{Prior distributions used in Case Study 1.\label{tab:C1_priors}}
    \begin{tabular}{ll}\toprule
    Prior & Distribution\\\midrule
    \PRIOR{k}          & \NORMAL{96.68}{100}    \\
    \PRIOR{c}          & \NORMAL{0.422}{0.1} \\
    \PRIOR{\sigma_f^2} & \NORMAL{0.5}{1}    \\
    \PRIOR{\ell}       & \NORMAL{0.1}{0.1}  \\
    \PRIOR{R}          & \NORMAL{0.1337}{1}  \\
    \bottomrule
    \end{tabular}
\end{table}

Using the standard Metropolis-Hastings accept reject kernel with a random walk proposal \cite{gelman2013bayesian}, samples are drawn from the Markov Chain until 20,000 are accepted and the first 2,000 of these are discarded as a ``burn-in''. %
Tuning the random walk proposals gave an acceptance rate of 17.9\% which is within reasonable bounds. %
From this set of samples it is possible to infer the posterior distributions in a Monte Carlo manner and histograms of these posteriors are shown in \cref{fig:duff_posteriors}. %
A certain degree of bias can be seen in the parameters of the linear system, both of which are overestimated when compared to the ground truth values known from the simulation. %
This is an interesting phenomenon which will be discussed fully later in the context of estimating the nonlinear component of the system from the estimated restoring forces.

\begin{figure}[h]
    \centering
    \includegraphics[width=\textwidth]{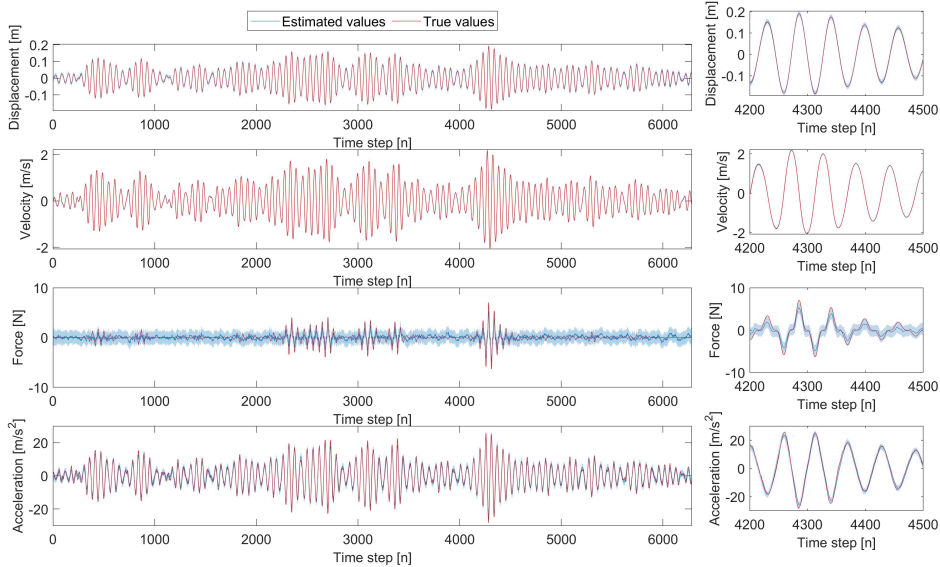}
    \caption{Top three frames GPLFM state estimates for the simulated Duffing model, on the right hand side sections are enlarged to see the quality of the estimates. Bottom frame shows reconstructed acceleration from the state estimates.}\label{fig:duff_states}
\end{figure}

Once the parameter posteriors have been estimated, it is also possible to estimate the distributions over the states in the model. %
It is worth commenting that these states represent the displacement and velocity of the oscillator as well as the unknown nonlinear restoring force. %
\cref{fig:duff_states} shows the estimation of these states obtained via Kalman filtering and RTS smoothing for the batch of data used in the identification alongside a reconstructed time series of the accelerations from those states. %
The uncertainty in the reconstructed acceleration is also shown through linear combination of the Gaussian uncertainty on all of the states.
Very high quality recovery of the first two states can be observed from the estimation, all data points lie within a $3\sigma$ confidence interval and if taking the expected values as point estimates, the error compared to the ground truth is 0.35\% for the displacement and 0.04\% for the velocity\footnote{To compute this error a normalised mean square error (NMSE) metric is used such that $J=100/N/\mathbb{V}[y]\sum_{i=1}^N\left(y_i-\hat{y}_i\right)^2$ where $y$ is the ground truth, $\hat{y}$ the estimated value and $\mathbb{V}[y]$ the variance of $y$ for $N$ data points.}. %
The associated error in the reconstructed acceleration is 2.81\% using the NMSE metric, which remains a very good score. %
The slight increase in relation to the displacement and velocity states is associated with the increased noise on the measured signal (as opposed to the noise free ground truth that the states are compared to) which is captured in the predictive uncertainty.


\begin{figure}[h]
    \centering
    \includegraphics[width=0.55\textwidth]{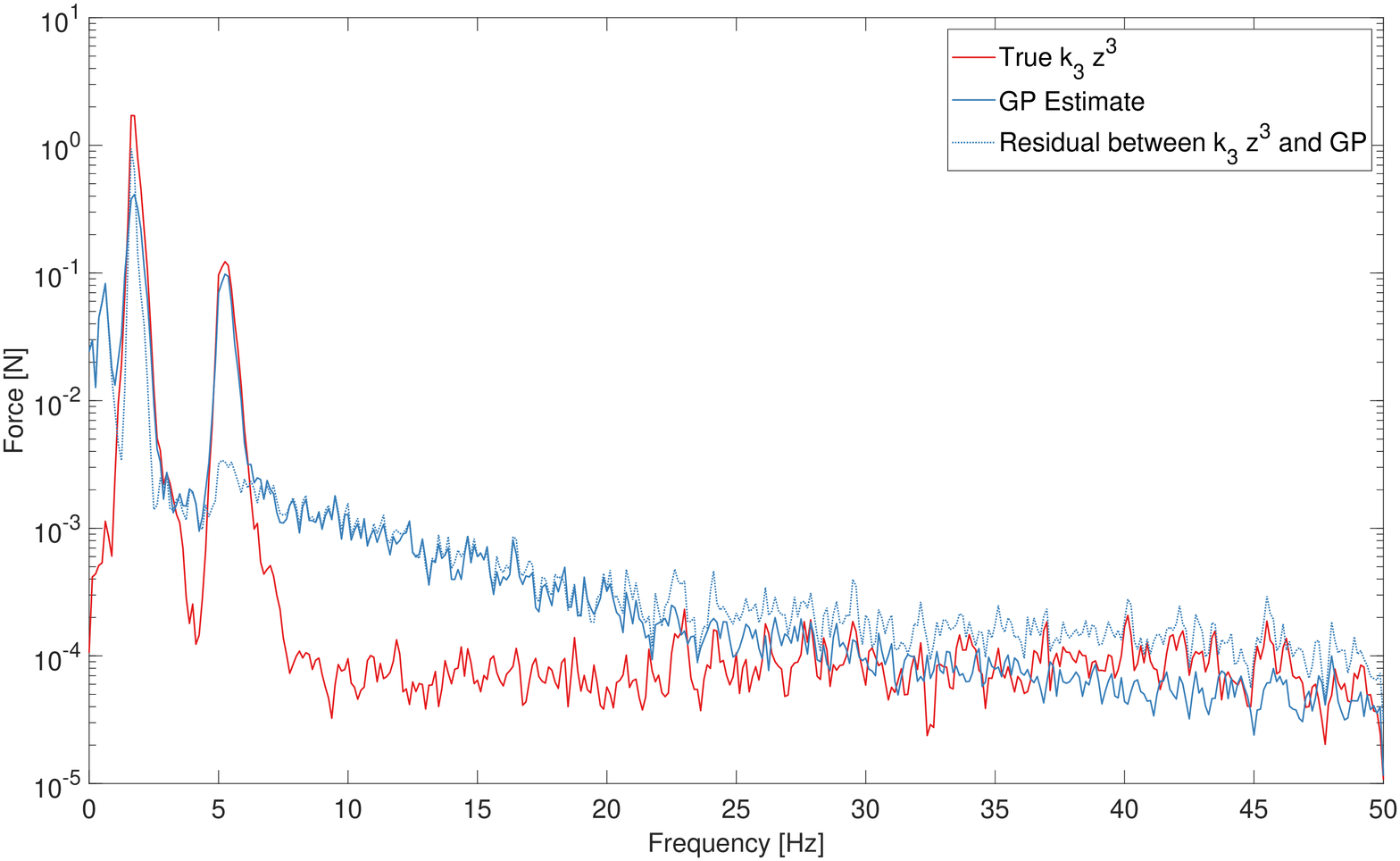}
    \caption{Frequency domain comparison of the third state related to the GP estimate.}\label{fig:x3_estimate}
\end{figure}

\begin{figure}[h]
    \centering
    \includegraphics[width=\textwidth]{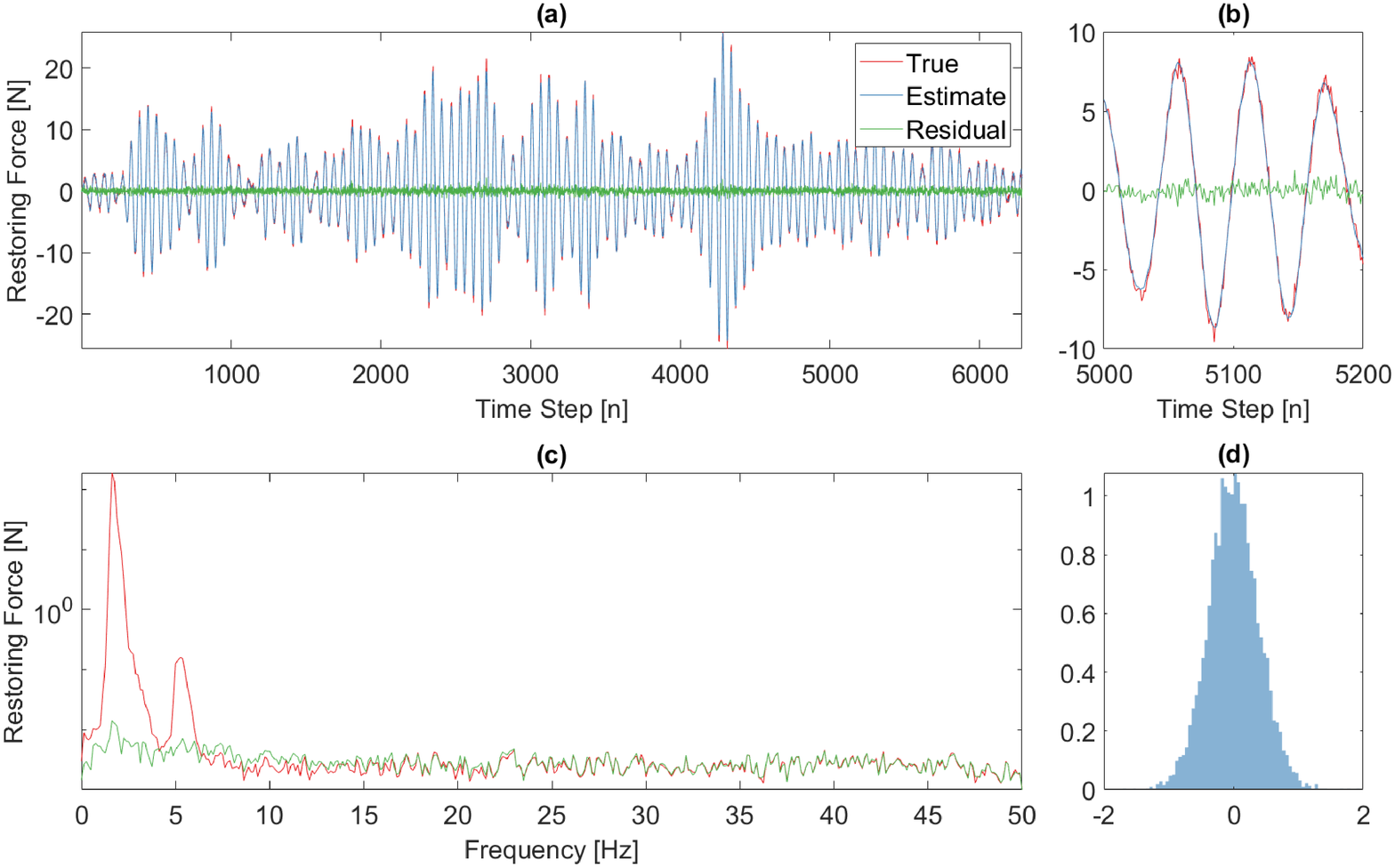}
    \caption{Estimation of the true restoring force in the model and its residuals. True values shown in red, estimates in blue and the residual in green. Panel (a) shows the time series; (b) a zoomed in section of that time series; (c) the frequency domain view of the restoring force and its residuals; (d) the distribution of residuals.}\label{fig:rf_estimate}
\end{figure}

In view of the goal of inferring the missing nonlinear component of the equation of motion, the second stage of the analysis is to extract samples of the unknown nonlinear restoring force. %
These can be obtained by sampling from the smoothing distribution of the constructed model, $p\left(\vec{x}_{1:T}\;\vert\;\vec{y}_{1:T},\vec{\theta}\right)$. %
These samples can be obtained in a backwards recursive manner based on the filtered system, see \cite{wills2012estimation}. %
The third estimated state in \cref{fig:duff_states} is that related to the GP which is being used to model the unknown nonlinear restoring force. %
Since this is a simulated system it is possible to compare this state with a ground truth as well. %
It can be seen that the fit of the nonlinear restoring force to the known ground truth is not as close to the ground truth as may be desired. %
The ground truth values of the nonlinear restoring force remain within the confidence bounds (three standard deviations) of the model, however, a significant increase in variance is observed and the expected value of this state does not lie as close to the ground truth as would be expected (60.3\% NMSE). %
A frequency domain view of the prediction and residual when compared to the ground truth \(k_3 z^3\) is shown in \cref{fig:x3_estimate}. %
Here the content of the difference between the GP estimate and the \(k_3 z^3\) term can be explored. %
It is seen that there is significant low frequency deviation in the GP close to 1\si{\hertz} and that there is increase high(er) frequency content from 7--25\si{\hertz}. %
The low frequency content is a result of the biased estimate of the linear stiffness which will be discussed shortly. %
The higher frequency deviation is a result of the relative roughness of the GP kernel prior as opposed to the polynomial nonlinearity, see \ref{sec:m32_results} where the choice of a kernel which is more times differentiable reduces this high frequency content. %
It is important to remember that the estimation of this state is not the final goal of the model, the aim of this procedure is to isolate the contribution of the nonlinear component such that it can be more easily identified and modelled, this will be expanded on shortly. %

Before showing how the estimated states may be used to learn the nonlinearity in the system, it is worth discussing further the bias in the linear system parameters and the difference between the estimated nonlinear restoring force (the GP state) and the known contribution. %
During the first stage of the estimation procedure, MCMC is employed to jointly infer the parameters of the model (including the linear stiffness and damping coefficients) and the states of the SSM. %
Considering the results shown in \cref{fig:duff_posteriors}, the damping coefficient $c$ is estimated accurately by the MCMC procedure, however, significant positive bias is observed in the linear stiffness $k$. 
The \emph{maximum a posteriori} (MAP) estimate of the stiffness is 107.38 $Nm^{-1}$ which is significantly higher than the ground truth of 100 $Nm^{-1}$ and counter to the perturbed prior which pulls that value lower. %
It will be seen that the total estimate of the restoring force is very accurate but this linear component is overestimated and the contribution of the GP is underestimated. %
This effect is not isolated to the current model under consideration, in fact is has been well established when combining GPs with physical models \cite{brynjarsdottir2014learning}.
The prior imposes a weak constraint on the values of the physical parameters which is informed by engineering insight into the system and in this case study the mass is assumed known which alleviates the most serious nonidentifiability problems. %
However, there is still a notable positive bias in the linear stiffness of approximately 7 \si{\newton\per\metre}, this increases the restoring force contribution of the linear stiffness term and conversely a lower contribution is made by the GP, leading to its difference from the ground truth. %
This phenomenon of bias in the physical parameters of a model when coupling it with a flexible machine learner is well established, for example see \cite{brynjarsdottir2014learning}, where even in simple examples the flexibility of the GP leads to a bias in the learnt physical parameters. %
In the case of a hardening stiffness nonlinearity in a dynamic system, it is intuitive that the linear stiffness may be overestimated to compensate for the contribution of the nonlinear terms, this is what is observed here. %

It is reasonable to question why, given the construction of the model, the third state in the system fails to properly capture the expected nonlinear restoring force $\hat{f}(z,\dot{z})$ which, for the Duffing system, should be simply $z^3$. %
To investigate, the estimate of the total restoring force is plotted in \cref{fig:rf_estimate}. %
This estimate is constructed by summing the estimated displacement multiplied by the MAP estimate of $k$, the estimated velocity multiplied by the MAP estimate of $c$ and the contribution of the GP which is the third state in the model. %
This can be compared to the known true restoring force in the model, $kz + c\dot{z} + k_3z^3$.
Considering this full restoring force as shown in \cref{fig:rf_estimate} (frame (a) and in detail in frame (b)), it can be seen that the estimate of the total restoring force is very accurate (0.34\% NMSE). %
The residuals of the estimate also appear close to white Gaussian noise indicating that there is no dynamic structure which has not been captured. %
This is further confirmed considering the frequency content of the signal (frame (c)) where the spectrum of the residuals is very close to flat and lies well below the spectrum of the true forcing close to its dominant frequencies. %
Additionally, considering the distribution of the residuals (frame (d)) a Gaussian shape is observed which is confirmed through a one-sample Kolmogorov-Smirnov test which fails to reject the null hypothesis at 0.1\% significance indicating that the samples are in fact Gaussian. %
Given this evidence, it is possible to conclude that the model has captured well the total restoring force present in the model, however, this reveals a point for further discussion regarding the quantity estimated by the Gaussian process.

\begin{figure}[h]
    \centering
    \includegraphics[width=0.8\textwidth]{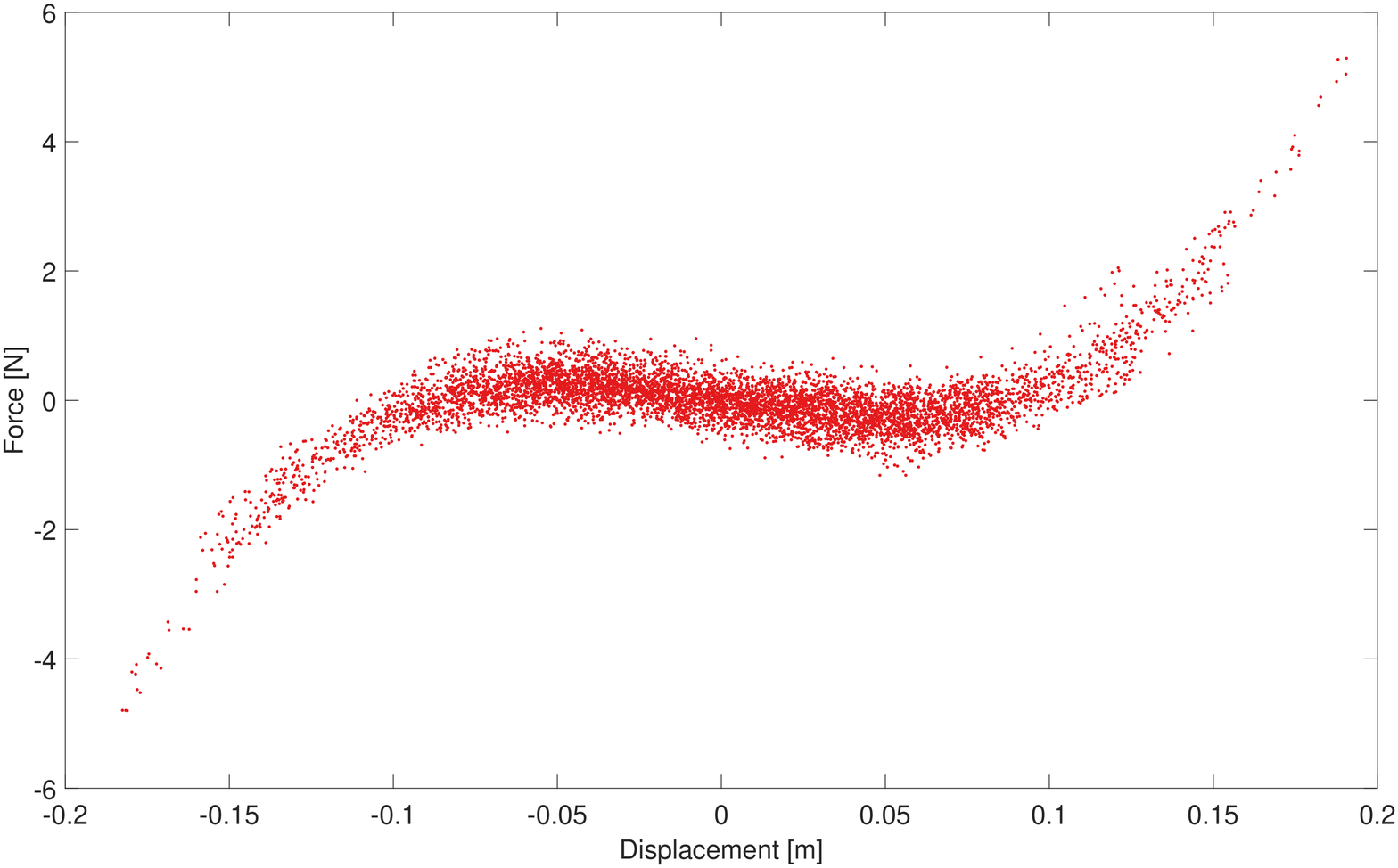}
    \caption{Plot of the nonlinear component of the model extracted from the GPLFM using estimated $z$ and $\hat{f}(z,\dot{z})$.}
    \label{fig:restoring_data}
\end{figure}

\begin{figure}[h]
    \centering
    \includegraphics[width=0.6\textwidth]{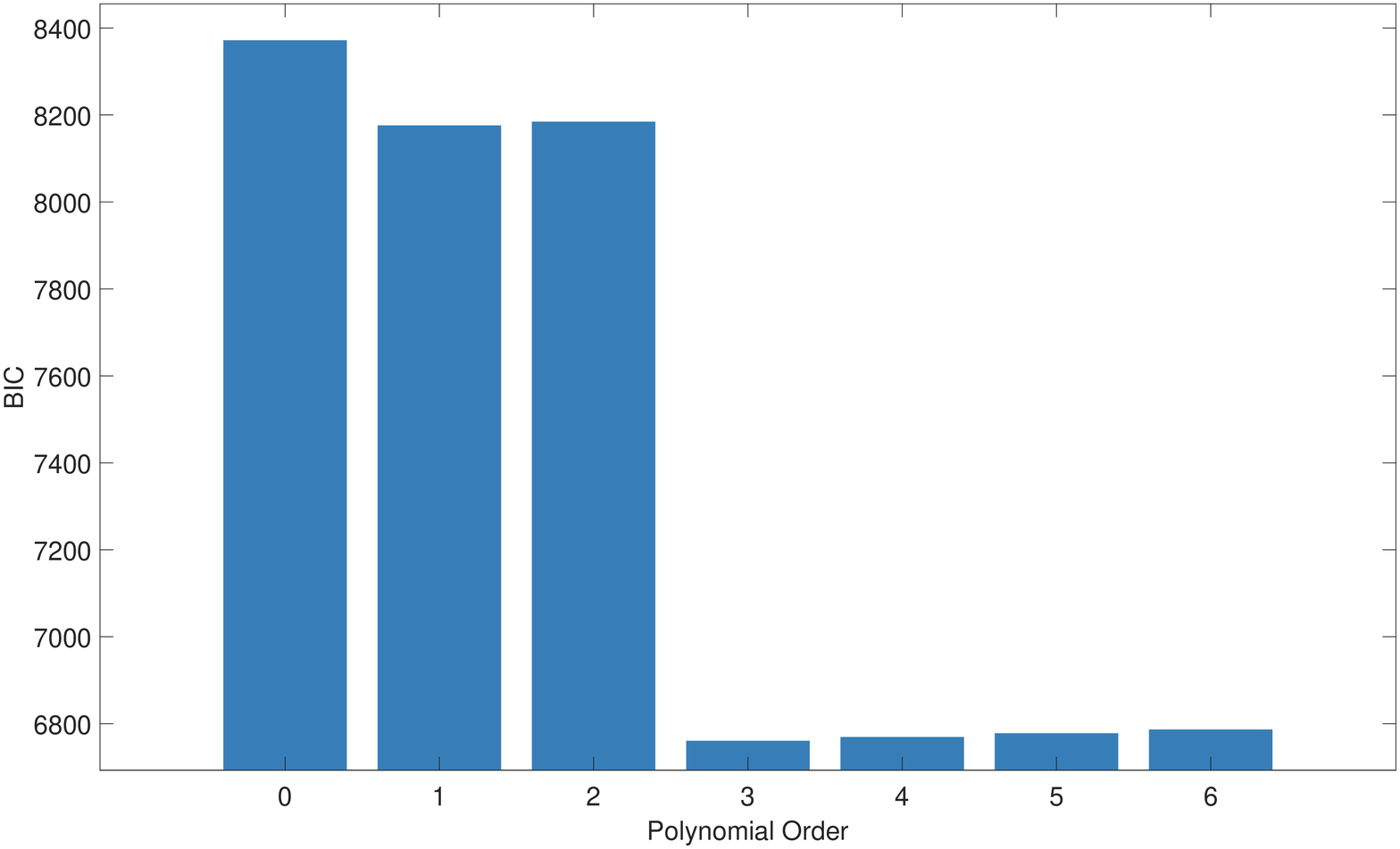}
    \caption{BIC for increasing polynomial model orders fitting the restoring force data.}
    \label{fig:BIC}
\end{figure}

Since it has been seen that the total restoring force in the system is well estimated in the GPLFM, the question remains of how to fit the nonlinear component of the system. %
As mentioned previously, there are a number of options for how to approach this task as it is close to the task in the classical restoring force surface method \cite{masri1979rfs}. %
For example, one could fit the extracted value from the GP component with another black-box model, e.g. a neural network (as in \cite{masri1993identification}) or another GP! %
However, one strength of the approach presented here as opposed to methods which simulate multiple nonlinear systems, e.g. the ABC approach of \cite{abdessalem2018model}, is that the GPLFM has estimated the latent components of the model, which are the unmeasured displacement, velocity and the (nonlinear) restoring force. %
Since the equation of motion is assumed in the form $m\ddot{z} + c\dot{z} + kz + \hat{f}(z,\dot{z}) = U(t)$ and the quantities for $z$, $\dot{z}$ and $\hat{f}(z,\dot{z})$ have now been estimated, the form of $\hat{f}(z,\dot{z})$ can now be plotted and inspected visually, which is not possible in other approaches. %
Using the estimated values for $z$ and $\hat{f}(z,\dot{z})$, the plot in \cref{fig:restoring_data} is produced. %
To learn the nonlinear system, one must fit this function which is a far easier task because it does not require running multiple nonlinear dynamic simulations. %
Importantly, the plot produced in \cref{fig:restoring_data} has been produced without any prior knowledge of the nonlinearity in the system, it is a result of the construction of the GPLFM. %

At this point it is necessary to determine which model to fit to the data in \cref{fig:restoring_data}. %
There are numerous approaches available, as discussed a purely black-box approach may be adopted, however, it may be at this point that engineering insight is able to add value. %
Having extracted the nonlinearity and in combination with insight into the physical structure an engineer may be able to make a reasonable judgement about the form of the nonlinearity, e.g. ``there is only a nonlinear stiffness component''. %
Of course, in the absence of certainty one could perform checks to confirm if the model suggested is in fact correct. %
This task is also greatly simplified since the function being learnt is now static. %
To demonstrate this a Bayesian Information Criterion (BIC) check over multiple polynomial model orders is performed on this data to confirm the visually suggested cubic shape. %
The BIC is calculated as: $$BIC = P\log N - 2\log \mathcal{L}$$ for $P$ parameters fitting $N$ data points and where the log likelihood of the fitted model is given by $\log \mathcal{L}$. %
Using a Bayesian linear regression the log likelihood of the model is available in closed form and the whole procedure can be conducted in under one second on a laptop computer. %
The lowest value of the BIC indicates the model with the ``optimal'' trade-off of model fit against model complexity. %
Taking the data extracted from the GPLFM and testing multiple polynomial models gives the results shown in \cref{fig:BIC}. %
It can be clearly seen that the BIC check confirms that this is indeed a cubic nonlinearity present in the system and identification can proceed. %
One additional factor which should be considered is the choice of the kernel used for the GP.
Since this choice only defines the prior over the restoring force the model should be relatively insensitive. %
In \ref{sec:m32_results} the same system is identified with a Mat\'ern 3/2 kernel and the results shown very similar performance to those with the Mat\'ern 1/2 kernel used in the results of this section.

\begin{figure}[h]
    \centering
    \includegraphics[width=0.8\textwidth]{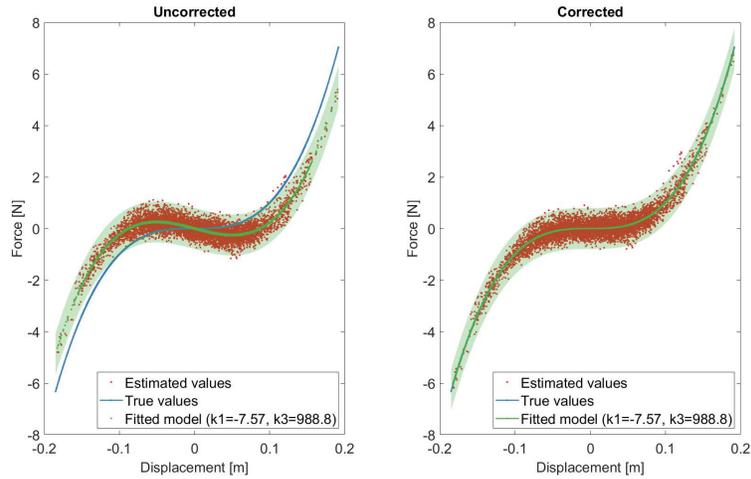}
    \caption{Estimating the cubic nonlinear term from the estimated states, the bias in the linear parameter from the GP is removed in the right-hand frame. The shaded area indicates the $3\sigma$ confidence intervals of the Bayesian linear regression.}
    \label{fig:duff_cubics}
\end{figure}

This procedure can now be seen in the context of the Duffing oscillator, where the nonlinear restoring force has been extracted as previously shown. %
Selecting the cubic model for the nonlinearity one can fit the data in \cref{fig:restoring_data} to obtain a nonlinear ODE which can be used in future simulations. %
This is done in \cref{fig:duff_cubics}. %

Remembering that the contribution of the GP does not perfectly capture the contribution of only the nonlinear term, the cubic function which is fit also contains the linear coefficient. %
Since the total restoring force $f(z,\dot{z}) = c\dot{z} + kz + \hat{f}(z,\dot{z})$ is estimated correctly (see \cref{fig:rf_estimate}) one can investigate the combination of the learnt $\hat{f}(z,\dot{z})$ with the previously identified linear parameters. %
Since the model being fit to the restoring force data is $\hat{f}(z,\dot{z}) = \alpha z + \beta z^3$, the learnt model may be substituted into $f(z,\dot{z})$ and the effect on the estimate of the linear coefficient is seen.
\begin{equation}
    \begin{aligned}
        f(z,\dot{z}) & = c\dot{z} + kz + \hat{f}(z,\dot{z})\\
        & = c\dot{z} + kz + \alpha z + \beta z^3\\
        & = c\dot{z} + (k + \alpha)z + \beta z^3
    \end{aligned}
    \label{eq:fix_duff}
\end{equation}
Therefore, given that the estimate of $f(z,\dot{z})$ is correct, the fit of the cubic to $\hat{f}(z,\dot{z})$ with a linear term in allows for the correction of the bias in the linear stiffness parameter identified in the MCMC procedure. %
This is shown visually in \cref{fig:duff_cubics} where the true $f(z,\dot{z}) = z^3$ term is shown in blue and the effect of correcting for the linear term is demonstrated.
It is worth noting that this approach will only work if the model fit to $\hat{f}(z,\dot{z})$ contains linear components which can be combined as in \cref{eq:fix_duff}. %

\begin{figure}[h]
    \centering
    \includegraphics[width=0.8\textwidth]{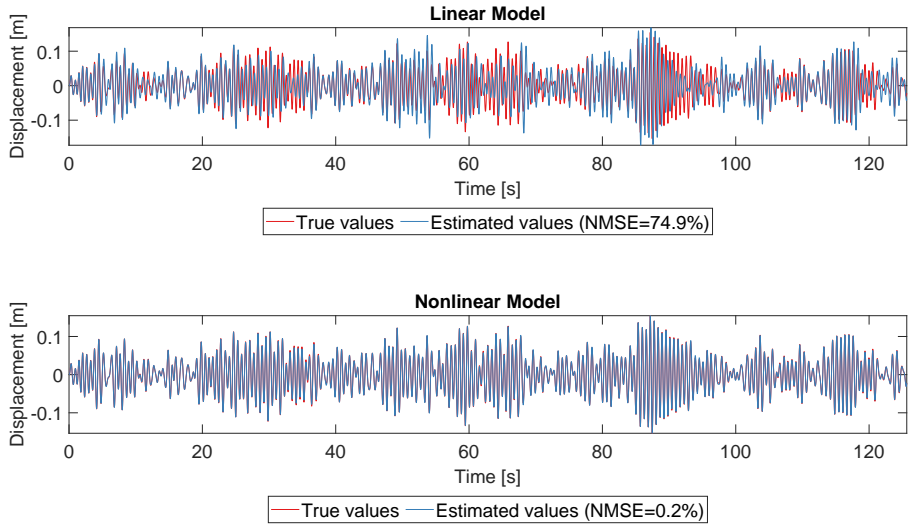}
    \caption{Comparison of simulation performance of the nonlinear model to that of an estimated linear model of the same system}
    \label{fig:duff_sim_performance}
\end{figure}

It is also possible to see the benefit of the proposed approach by considering the performance of the learnt model when used to simulate the system that was identified. %
Given the identified models, using the expected values of the parameters, with the nonlinear component as learnt in \cref{fig:duff_cubics}, the response of the system can be simulated.
This is shown in \cref{fig:duff_sim_performance} for both the linear model and the nonlinear model which have been fitted. %
The performance of the linear model, i.e. only the dynamic model learnt in the MCMC when the GP is removed, is seen to be significantly lower than the estimated nonlinear model. %
When using the identified linear model, the NMSE is 74.9\% a level which for almost all applications would be considered unsatisfactory. %
However, once the nonlinearity has been identified and included in the model used for simulation this NMSE falls to 0.2\%. %
This represents a significant increase in quality of fit, which is to be expected. %
Having seen that the proposed methodology is effective on a simulated dataset, it can now be shown on an experimental benchmark.

\subsection{Silverbox Benchmark}
\label{sec:silverbox}

The Silverbox benchmark \cite{wigren2013three} is an experimental dataset produced by an electrical system which implements a nonlinearity, similar in behaviour to the theoretical Duffing oscillator. %
It has been investigated previously in the literature as a benchmark dataset for nonlinear system identification, for example see \cite{marconato2012identification,schoukens2018nonlinear,worden2018evolutionary,kocijanparameter}. %
The benchmark itself aims to replicate, electronically, the behaviour of a Duffing oscillator seen previously in \cref{eq:duffing}. %
Although it may be the case that it does not perfectly produce such a response it is a very good approximation and serves as a challenging benchmark. %
Two datasets are available from the benchmark but one, referred to as the `arrowhead' is the most commonly used in identification and testing. %
Two excitation signals are applied, first is a filtered Gaussian white noise with an increasing amplitude (40,000 samples) which forms the `head' of the arrow then ten realisations of an odd random phase multisine of length 8192 samples at a fixed amplitude. %
See \cite{wigren2013three} for the complete details.

\begin{figure}[h]
    \centering
    \includegraphics[width=0.6\textwidth]{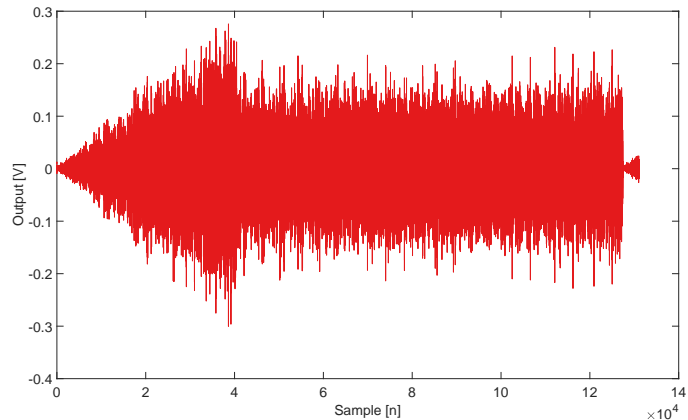}
    \caption{The measured `arrowhead' response of the Silverbox benchmark data.}
    \label{fig:arrowhead}
\end{figure}

In keeping with most approaches seen in the literature, this work uses the `tail' of the data to identify a nonlinear model of the system which is then tested on the `head' of the arrow (points 1 to 40,500). %
The identification is based on samples from within one realisation of the odd random phase multi-sine in the tail of the arrowhead, specifically data points 49,278 to 52,350 are used in the training phase (a total of 3072 observations). %
In training and testing these data were upsampled by a factor of four using a cubic spline interpolation. %
When reporting testing errors, the simulated data are downsampled and compared against the original collected corresponding time points.
A GP with a Mat\'ern 1/2 kernel is used to estimate the unknown nonlinear restoring force as a function in time.
Its hyperparameters $\sigma_f^2$ and $\ell$ are estimated along with the parameters of the linear dynamic system by means of an MCMC inference over the linear SSM. 

\begin{figure}[h]
    \centering
    \includegraphics[width=0.8\textwidth]{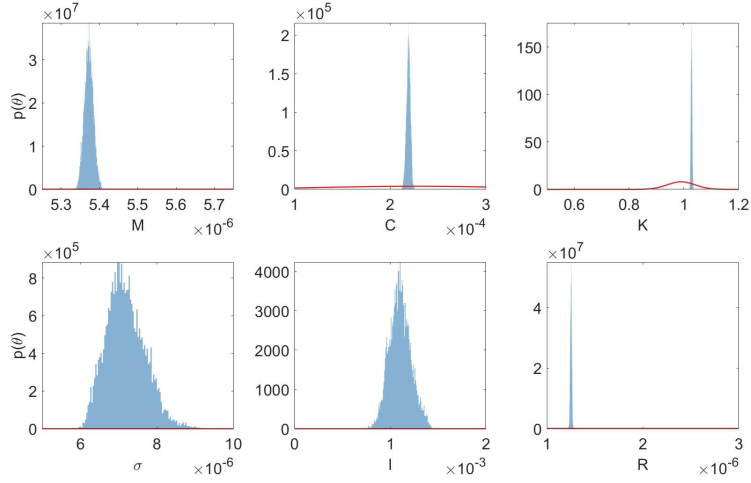}
    \caption{Parameter posteriors from the MCMC inference over the state-space model for the Silverbox. Solid (red) lines indicate the probability density function of the prior and the histograms (blue) are built from samples of the posterior.}
    \label{fig:silverbox_posteriors}
\end{figure}

Priors over the parameters were set on the basis of the results reported in \cite{kocijanparameter}. %
The priors were defined as in \cref{tab:C2_priors}, these priors are established on the basis of the best engineering knowledge of the system which would be available in application of the method. %
To explore the sensitivity of the posteriors to the choice of prior, further results are shown in \ref{sec:silverbox_priors}.
Given this specification of the priors and using the likelihood of the linear state-space model a Metropolis-Hastings based MCMC inference was run. %
Ten thousand samples are drawn from the Markov Chain of which two thousand are discarded. %
From these samples the posterior distributions over the system parameters and the hyperparameters of the GP are recovered and shown in \cref{fig:silverbox_posteriors}. %
Since this is a physical system, there are no ground truth values to be compared to but it can be seen that the posterior distributions have not shifted too far from the specified priors and that there has been a decrease in the variance of the estimates.
The real test of this identification will be in the simulation of the test data shown later.

\begin{table}[h]
    \caption{Prior distributions used in the Silverbox case study.\label{tab:C2_priors}}
    \centering
    \begin{tabular}{cc}\toprule
    Prior & Distribution\\\midrule
    \PRIOR{m}          & \NORMAL{5.3732\times 10^{-6}}{1\times 10^{-5}}    \\
    \PRIOR{c}          & \NORMAL{2.2653\times 10^{-4}}{1\times 10^{-4}}    \\
    \PRIOR{k}          & \NORMAL{0.99}{0.05}                               \\
    \PRIOR{\sigma_f^2} & \NORMAL{0.005}{0.05}                              \\
    \PRIOR{\ell}       & \NORMAL{0.4}{0.05}                                \\
    \PRIOR{R}          & \NORMAL{1.252\times 10^{-6}}{5\times 10^{-6}}     \\
    \bottomrule
    \end{tabular}
\end{table}

\begin{figure}[h]
    \centering
    \includegraphics[width=0.9\textwidth]{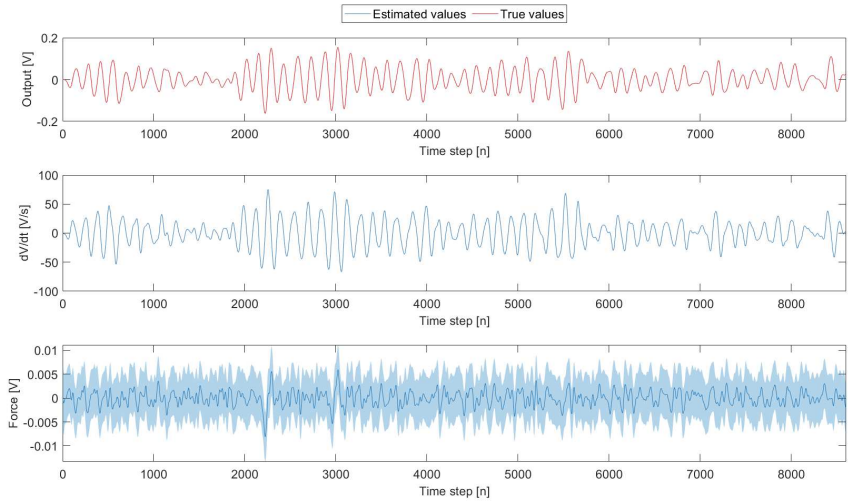}
    \caption{GPLFM state estimates for the Silverbox. Note that ground truth values are not available for $dV/dt$ or the nonlinear restoring force state.}
    \label{fig:silverbox_states}
\end{figure}

The estimates of the system parameters also allow samples of the state trajectories to be generated as before by sampling from the smoothing distribution of the LGSSM. %
Since only the output voltage of the system, corresponding to the displacement in a mechanical oscillator, is collected, the ground truth is only available for that state. %
It should be noted that this also affected the observation model used in the system, leading to $\mat{C} = \begin{bmatrix}
    1 & 0 & 0
\end{bmatrix}$ and $\mat{D}=0$ when constructing the state space model. %
It can be seen in \cref{fig:silverbox_states} that there is very good agreement in this state. %
The estimates of the voltage (output) state and the first derivate ($dV/dt$) states show very low levels of uncertainty in the estimates and the accuracy of the first state would lead to the reasonable assumption that $dV/dt$ has also been well captured. %
This assumption is additionally supported by the results seen in the previous numerical case study. %
With respect to the forcing state which relates to the input voltage, the degree of uncertainty is significantly higher as was seen in the previous case. %
There are also periods as seen before where in regions of high displacement clear spikes in the nonlinear restoring force can be seen which may be related to increased effects of the nonlinear component in the system. %

\begin{figure}[h]
    \centering
    \includegraphics[width=0.7\textwidth]{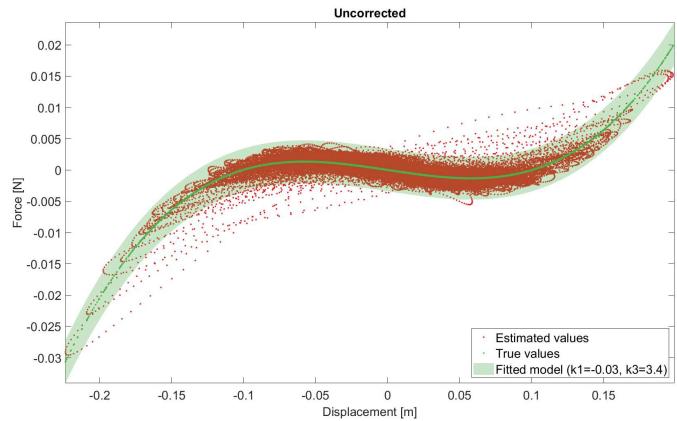}
    \caption{Fitting of the cubic parametric nonlinearity to the samples of the nonlinear restoring force}
    \label{fig:silverbox_cubics}
\end{figure}

Using these samples it is possible to visualise and construct a regression which models the nonlinearity in the system. %
Since some knowledge regarding the nonlinearity is known \emph{a priori} in the case of the Silverbox dataset it was possible to specify a parametric form of nonlinearity to be fit based on these samples of the states. %
That parametric form is one of a cubic polynomial with respect to the displacement state. %
It is assumed that none of the damping behaviour of the system has been transferred into the third state which relates to the GP model of the nonlinear restoring force. %
The samples can be drawn from the GP in time which is now the estimate of the missing contribution of the nonlinear restoring force, given the assumption of a parametric cubic nonlinearity the regression in \cref{fig:silverbox_cubics} can be learnt. %
As before, a Bayesian linear regression is used to additionally estimate the uncertainty in the estimate. %
Of interest in this result is that the residuals of the model are clearly not Gaussian i.i.d., this would suggest that there remains, at least in the samples, some dynamic behaviour not captured by the chosen parametric model of the nonlinearity and this could be another potential source of bias in the parameters. %
A subject of further investigation is determining whether this is an artefact of the methodology proposed or a result of the approximation of the Duffing behaviour through its realisation in an electronic form.

Despite the discrepancies seen between the fitted model and the samples of the states, for a large proportion of the data the fitted model is a good representation of the behaviour. %
In possession of this nonlinear model it is also now possible to consider the performance of the overall identification approach through simulation of the test data. %
The linear parameters of the model can be taken as the MAP estimates of the posterior distributions in \cref{fig:silverbox_posteriors} and the parameters of the cubic model learnt from samples of the states are added to this to establish the nonlinear model, the values used in simulation are shown in \cref{tab:map_silverbox}. %
Given the description of the cubic ODE, the inputs for the ramp up at the beginning of the arrowhead are used to simulate the response.

\begin{figure}[h]
    \centering
    \includegraphics[width=\textwidth]{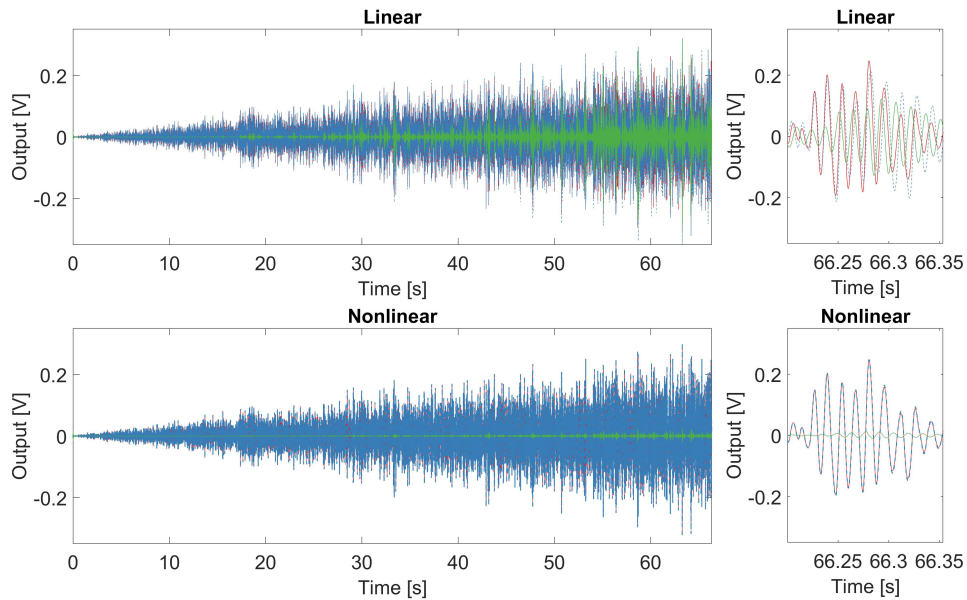}
    \caption{Testing portion of the `arrowhead' signal. Ground truth is shown in red, simulated response in dashed blue and the residual in green. The final 750 points of test data are enlarged to the right hand side of the figure for more detailed comparison.}
    \label{fig:arrowhead_est}
\end{figure}

\begin{table}[h]
    \caption{MAP estimates of paramters used for test simulation.\label{tab:map_silverbox}}
    \centering
    \begin{tabular}{cc}\toprule
    Parameter & Value\\\midrule
    $m$          &  5.3722 $\times 10^{-6}$  \\
    $c$          &  2.1905 $\times 10^{-4}$  \\
    $k$          &  0.9932  \\
    $k_3$        &  3.4239  \\
    \bottomrule
    \end{tabular}
\end{table}

For comparison, the simulation performance \cite{ljung1998system} of the model is compared for the learnt linear and nonlinear models in \cref{fig:arrowhead_est}\footnote{Note, that owing to the good performance in large regions for both models the ground truth (red) is overlapped by the simulated signal (blue), the residuals (green) make clear the performance increase from the linear to the nonlinear model.}. %
Starting from assumed zero initial conditions, the models attempt to recreate the previously observed behaviour of the Silverbox with access only to the inputs to the system. %
Using only the learnt linear model, the results in the top frame of \cref{fig:arrowhead_est} are obtained. %
With the ground truth in red and simulation in blue it can be see that the linear model does quite well especially in the initial portion of the data, roughly the first 20 seconds. %
In this lower amplitude excitation region the effect of the nonlinearity in the system is not as pronounced and the methodology adopted appears to have effectively captured the underlying linear subsystem. %
However, as the input amplitude has increased the error in this linear prediction grows considerably as can be seen by the large increase in the residual of the model, shown in green. %
The nonlinear model on the other hand (lower frame of \cref{fig:arrowhead_est}) preserves very good performance at low excitation levels, in fact showing even lower residuals in the initial portion of the data, as well as retaining excellent performance as the input magnitude increases. %
There is still an increase in the residual magnitude as the ramp continues, especially when the level of input exceeds that seen in the training phase of the methodology. %
This is likely due to some bias in the identified parameters of the nonlinear model or as a result of some unmodelled nonlinearity in the system. %
It is reassuring, however, that very good performance is maintained by the identified nonlinear model even in `extrapolation', in the sense of the input magnitude level. %
For some quantitative comparison the normalised mean squared errors are computed for both the linear and nonlinear model as before. %
The resulting errors are 18.47\% for the linear model and 0.17\% for the nonlinear model, the normalised error of 0.17\% corresponds to a root-mean-squared-error of 0.0021 in the testing data which is competitive with results seen in the literature, e.g.\ \citet{schoukens2018nonlinear}.

\begin{figure}[h]
    \centering
    \includegraphics[width=0.8\textwidth]{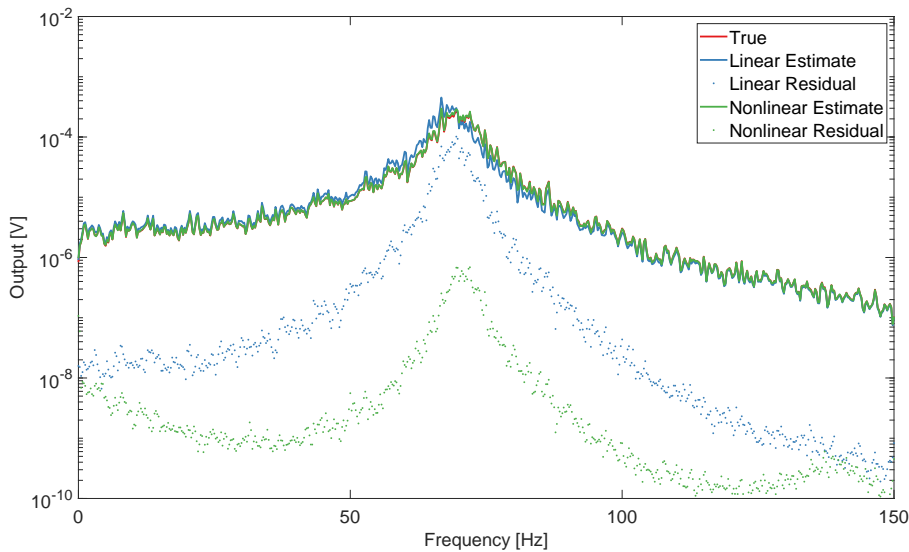}
    \caption{Frequency domain comparison of the arrowhead fit.}
    \label{fig:freq_arrowhead_est}
\end{figure}

In addition to the time series comparison of the linear and nonlinear simulation of the system it is also possible to explore the frequency content of the fits and their relative quality. %
This comparison is shown in \cref{fig:freq_arrowhead_est}. %
While both models appear to have relatively well captured the frequency content of the arrowhead, the plots of the residuals show the significant improvement which is observed when moving to the identified nonlinear model. %
This is consistent with the results seen in the time series and also indicates that the most significant benefit from the nonlinear model is seen near the resonance of the system which is to be expected.

In summary, the proposed approach for inference over the nonlinear restoring force in the system and the linear system parameters has proven to be very effective in identification of the Silverbox. %
The use of MCMC to learn the underlying linear system and generate samples of the potential nonlinear restoring force as well as the unobserved states of the system has allowed a nonlinear model of the dynamics to be learnt which was proven to perform well on an independent test set in simulation.

\section{Discussion}

Having seen the performance of the method, it is now possible to make some observations and comments on the proposed approach. %
It will be possible to highlight the potential benefits of the GPLFM viewpoint but also present some of the potential challenges in applying this method as it currently stands. %
Those challenges should be seen to inform avenues for further investigation. %
The authors are also mindful that, while not formally proved, it is reasonable to assume that there is \emph{no free lunch} in system identification, as is famously the case in optimisation \cite{wolpert1997no}. %
In view of this, the authors will not (and should not) claim that the GPLFM approach presented in this work supersedes other efforts in the system identification field for all models but do argue its utility for certain classes of problem. %

At the core of the method is the idea to break down the complex nonlinear system identification problem into two (hopefully) simpler tasks:
\begin{enumerate}
    \item To identify the displacement and velocity of a system alongside some additional unknown forcing represented by the GP
    \item To use these identified quantities to fit a nonlinear model in a \emph{static} manner
\end{enumerate}

It has been seen that the GPLFM, in its state-space form, provides a model with sufficient flexibility to achieve the first task. %
The estimates of the displacement and velocity, which can be traditionally challenging in the case of noisy data, appear well estimated in the first case study where they can be compared to the ground truth. %
Of note, as discussed, is the bias in the estimated linear parameters of the system in this initial stage of fitting. %
However, the value of this initial stage, despite the linear parameter bias, seen is that the identification problem is translated from one which is inherently dynamic to one which is now static, i.e.\ all the component terms in the ODE are now known and \(f(z,\dot{z})\) can be fitted directly and, by including linear terms, the linear parameter bias can be corrected.

Contrasting the proposed approach to a more common nonlinear system identification approach, e.g.\ a deterministic optimisation of a nonlinear model, e.g.\ \cite{worden2018evolutionary}, or a Bayesian technique such as ABC-SMC \cite{abdessalem2018model}, there are two advantages to the methodology in this work. %
Firstly, during the first stage of the identification (recovering the restoring force) the modeller is not required to suggest any candidate nonlinear system models, instead the restriction in the current work is that the system should be a single-degree-of-freedom, be representable by \cref{eq:eom_rflfm} and that the unknown contribution of the nonlinearity must be capable of being modelled by a GP in time with a stationary kernel. %
Addressing the first limitation is the most pressing within this framework, although the authors are hopeful that this will be the subject of future work in the near future.
The main challenge is to address identifiability problems with how the latent forces enter the multiple-degree-of-freedom system so as to maintain physical interpretability. %
The second limitation on the method is found to be very small when considering the class of nonlinear problems encountered in mechanical engineering where it is generally assumed that systems exhibit second order behaviour (in the single-degree-of-freedom case). %
Finally, the requirement for a stationary kernel in the GP is also a relatively weak constraint on the problem, one point of interest for the modeller is to consider the ability of the GP to model the force if there is a discontinuity in the restoring force \emph{time series}, bearing in mind that GP will still attempt to model the signal but will smooth the ``sharp corners'' in the signal.

The second advantage of the proposed GPLFM approach is in its significant computational efficiency as compared to optimisation of a nonlinear dynamic model \cite{worden2018evolutionary}, a procedure such as ABC-SMC \cite{abdessalem2018model} or even MCMC over a nonlinear state-space model. %
The source of this advantage is in the retention of a linear state-space model despite the nonlinearity in the system, where the mismatch between the linear dynamic prior (GP) and the nonlinear contribution is corrected through the posterior estimation of the states from the smoother. %
This means that computationally, the estimation can be carried out in \emph{linear time} (\(\mathcal{O}(T)\)) as opposed to in a simulation based approach (optimisation or ABC, including its more efficient variants \cite{abdessalem2019model}) which requires multiple time series simulations (i.e.\ Runge-Kutta or similar integration approaches) or a method which required estimation of the smoothing distribution of a nonlinear state-space model. %
Once the states and missing contribution have been recovered in this computationally efficient manner the remaining fitting of the nonlinearity happens statically, that is without the need to simulate time series responses. %
For example, in the cases shown here (with the choice of a polynomial basis for the nonlinearity) the fitting of the nonlinearity reduces to linear regression. %
This fitting of the nonlinearity will be much faster in the static setting than the dynamic one and allows the modeller to interrogate this static relationship graphically, providing physical insight into the form of the nonlinearity before fitting begins which can guide the class of models to be tested. %
Even if additional engineering insight cannot be embedded at this stage, the faster fitting of this static function still enables more efficient model selection, demonstrated here was the use of BIC although more sophisticated approaches are also possible.

It is worth considering then when the approach proposed in this work may not be recommended by the authors. %
In its current form the framework has not yet been extended to multiple-degree-of-freedom systems, therefore, unless each degree-of-freedom can be isolated in testing the method is not applicable. %
Secondly, the modeller will encounter difficulty when the nonlinear term in the equation of motion is not readily fit in the second stage. %
This situation will occur, for example, if there is a \emph{dynamic nonlinearity} in the system, e.g.\ hysteresis, as opposed to a static functional form for \(f(z,\dot{z})\), e.g. polynomial in \(z\) and/or \(\dot{z}\). %
If in a situation where this form of nonlinearity is expected, it may still be useful to complete the first stage in the proposed identification framework, extracting the restoring force, to inspect it and confirm that a static function is not appropriate. %
In future it would be an interesting avenue of investigation to consider if the extracted restoring force could be fit in the second stage of the GPLFM process by means of fitting another ODE to that signal. %
At that point, however, the difficulty in identification may not be significantly reduced and a one-step approach such as the ones discussed may be more appropriate, especially if a range of candidate models for the system are readily available. %
Despite these potential limitations of the method, the authors would propose that the presented GPLFM approach should be adopted as another available tool for the modeller which would excel when there is little \emph{a priori} knowledge regarding the nonlinearity (only that it is static if wanting to fit the data directly in the second step) and which provides a comparatively fast, interpretable approach to nonlinear system identification.

\section{Conclusion}
\label{sec:conclusion}

This paper has proposed a methodology for identification of nonlinear dynamic systems when the form of the nonlinearity is not known \emph{a priori} and when the measurements may be noisy. %
The identification procedure requires two steps. %
The first step is concerned with extracting the contribution of the nonlinear component of the system, modelling the system as a linear oscillator with an additional force acting on it, in the form of a GP in time. %
The second step, which has not been the focus of this paper, is to build a model based on that contribution, although it is shown in the first case study how given the extracted data model selection techniques can be used to fit the nonlinearity. 

In order to extract the action of the nonlinear component of the model, the nonlinear restoring force, the approach proposed in this work models that force as a Gaussian process in time. %
Through Bayesian filtering and smoothing, this GP prior is updated to give an estimate over the possible contributions from the nonlinear term. %
Given this estimates of the internal states of the system and of the `missing' nonlinear component can be sampled and used to construct the nonlinear model. %
This nonlinear model need not be known \emph{a priori} since it is fit to the samples of the nonlinear restoring force which are extracted nonparametrically from the system since they are modelled by the Gaussian process. %
Once in possession of those samples, the practitioner may employ a wide range of modelling techniques, from expert intuition regarding the parametric form of the nonlinearity to fully \emph{black-box} modelling of those samples.
It was discussed how the inclusion of the linear terms in the initial identification of the model can introduce bias into the true linear parameters of the ODE. %
It was additionally shown that the estimation of the total restoring force, linear and nonlinear, remains very accurate despite this bias. %
The model for the nonlinearity could then be learnt either solely from the GP component as was shown in this work or using this estimated total restoring force. %
When the learnt nonlinearity from the GP included a linear term related to the displacement, this could be used to correct the bias in the estimated linear parameters.

The case studies in this paper have then shown that it is possible to realise such an identification procedure. %
Very good results have been shown both on simulated and experimental datasets.
However, a number of future avenues of research remain open. %
The first of these is to address more fully the second step of the identification procedure when a parametric form for the nonlinearity is not easily described or available. %
Some example has been given in this work where a parametric identification is performed with an number of polynomial models as candidates. %
BIC confirmed that a cubic stiffness model was an appropriate model in this case, however, in the future a fully nonparametric approach could also be explore or a more sophisticated model selection regime.
The second future challenge which should be addressed is the extension of this methodology to a multi-degree-of-freedom case where the nonlinear restoring forces may act on more than one degree of freedom in the model. %
One example of this may be a two degree-of-freedom lumped mass system where the masses are connected by a nonlinear spring (e.g.\ cubic). %

In conclusion, this work has introduced a methodology for identification of nonlinear dynamic systems. %
The core of the approach has been to separate the contribution of the nonlinear terms from the underlying nonlinear system by application of a Gaussian process latent force approach. %
The success of this method in the case studies shown would suggest that further investigation into approaches of this type could be a valuable pursuit in the future.

\section*{Acknowledgements}
The authors are grateful for the assistance of E.J.\ Cross in proof reading this manuscript and for many interesting conversations related to the work.
This research was supported by funds from the Centre for Oil and Gas - DTU/Danish Hydrocarbon
Research and Technology Centre (DHRTC), in support of T.\ Friis, as a visiting student in Sheffield. T.J.\ Rogers is supported by the University of Sheffield.

\bibliographystyle{unsrtnatemph}
\bibliography{refs}

\newpage
\appendix
\section{Alternative Kernel Results}\label{sec:m32_results}
\setcounter{figure}{0}  

The same approach as seen applied to the Duffing oscillator in the main body of the text is applied to the same system with the same forcing, only varying the use of a Mat\'ern 3/2 kernel as opposed to the 1/2 kernel used in the first study.

\begin{figure}[H]
    \centering
    \includegraphics[width=0.8\textwidth]{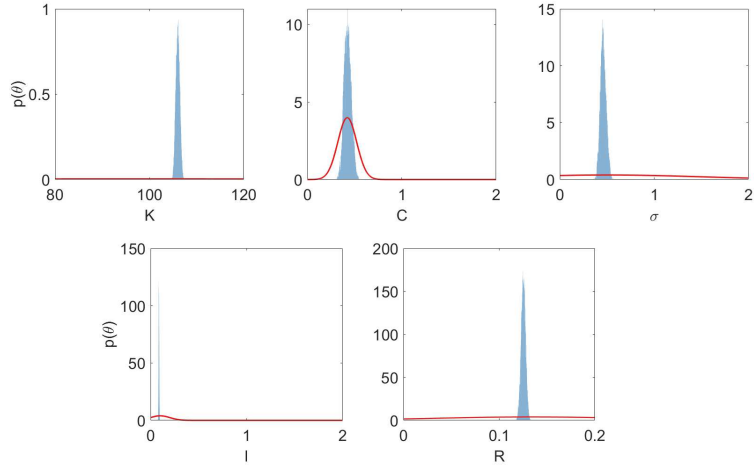}
    \caption{Parameter posteriors from the MCMC inference over the state-space model with a Mat\'ern 3/2 kernel.}
    \label{fig:duff_posteriors_m32}
\end{figure}

The results of the MCMC run over the system shown in \cref{fig:duff_posteriors_m32} show that the posteriors lie very close to those observed in the main body. %
Since the change of kernel does not affect the number of hyperparameters in the GP, the number of parameters to be learnt via the MCMC procedure remains the same. %

\begin{figure}[H]
    \centering
    \includegraphics[width=\textwidth]{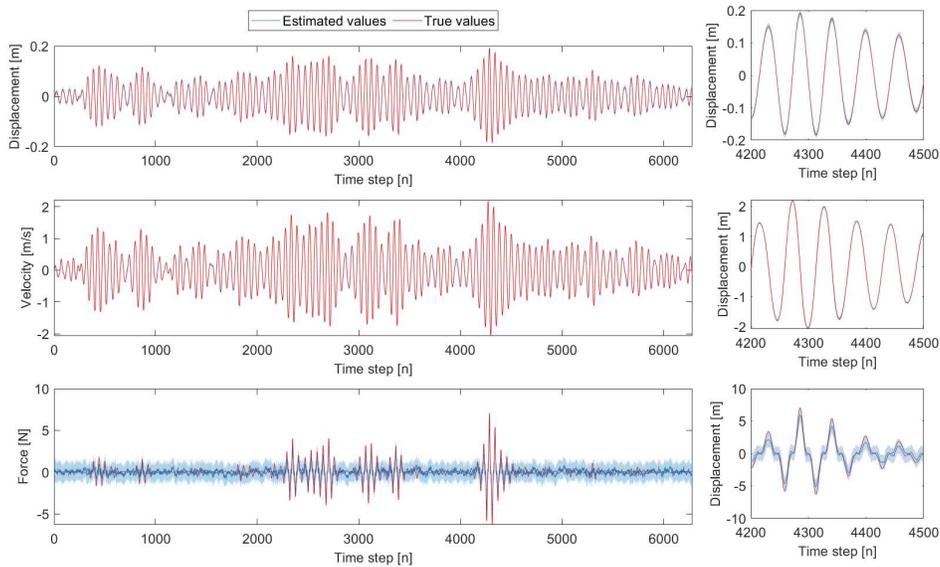}
    \caption{GPLFM state estimates for the simulated Duffing model, with a Mat\'ern 3/2 kernel, on the right hand side sections are enlarged to see the quality of the estimates.}\label{fig:duff_states_m32}
\end{figure}

The estimation of the states with the Mat\'ern 3/2 kernel is shown in \cref{fig:duff_states_m32} for the physically meaningful states of the model. %
The fourth state corresponding to the derivative of the force with respect to time is not shown as it is not simple to interpret its contribution to the model. %
As can be seen, the issues with overestimation of the linear stiffness remain and the estimate of the forcing state shows some continued underestimation. %
The quality of the fit in the displacement and velocity states is comparable to that seen when using the Mat\'ern 1/2 kernel. %
As may be expected, in the forcing state estimated by the kernel there is a less pronounced high frequency component in the estimate. %
This can be attributed to the less rough nature of the Mat\'ern 3/2 kernel. %

\begin{figure}[H]
    \centering
    \includegraphics[width=\textwidth]{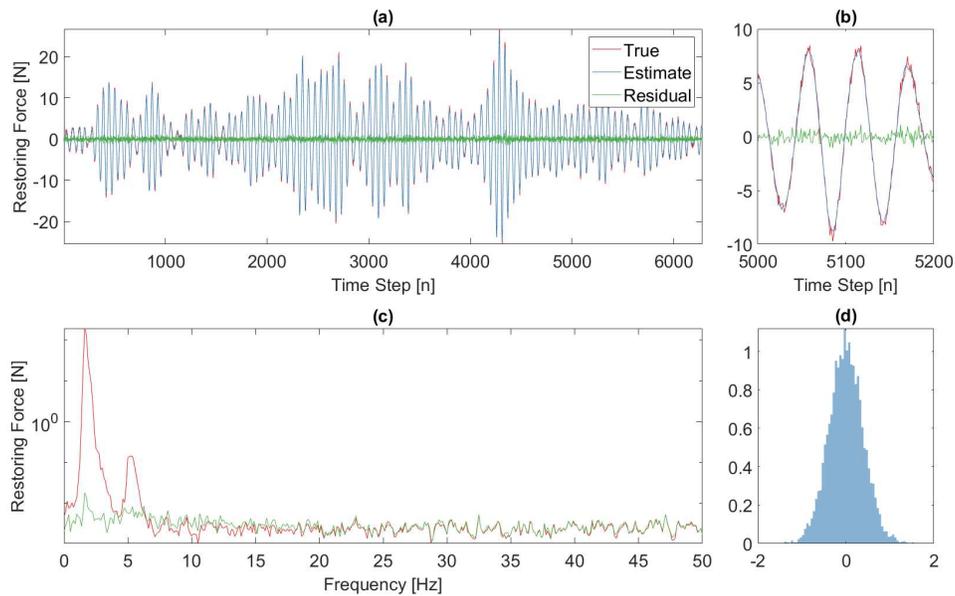}
    \caption{Estimation of the true restoring force in the model and its residuals when using the Mat\'ern 3/2 kernel. True values shown in red, estimates in blue and the residual in green. Panel (a) shows the time series; (b) a zoomed in section of that time series; (c) the frequency domain view of the restoring force and its residuals; (d) the distribution of residuals.}\label{fig:rf_estimate_m32}
\end{figure}

Turning attention again to the estimation of the full restoring force, this is shown in \cref{fig:rf_estimate_m32}. %
The quality of the fit on this quantity is again very similar to that observed in the main body. %
If the NMSE of the estimation is considered it is slightly larger at 0.3\% as opposed to 0.2\% with the Mat\'ern 1/2 kernel. %
However, both of these are excellent fits and it would be hard to attribute this change in NMSE to a drop in performance when changing the GP kernel. %

\begin{figure}[H]
    \centering
    \includegraphics[width=0.8\textwidth]{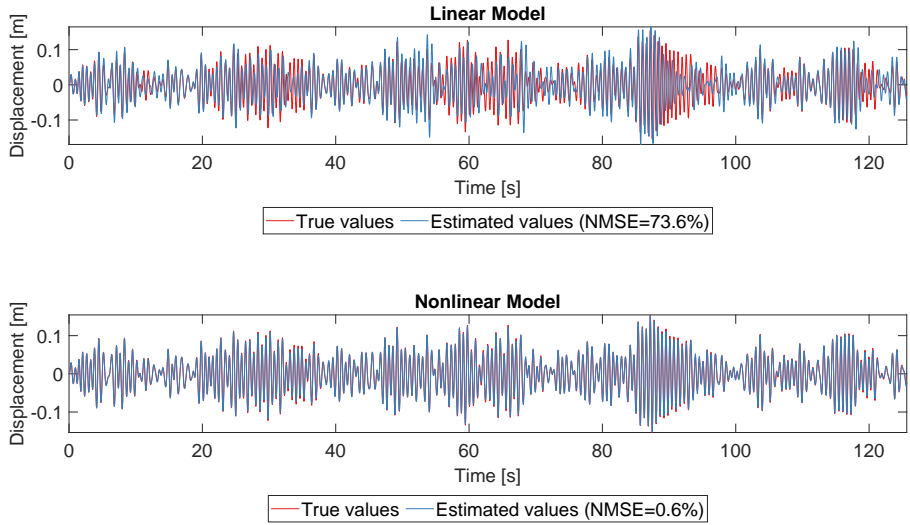}
    \caption{Comparison of simulation performance of the nonlinear model to that of an estimated linear model of the same system}
    \label{fig:duff_sim_performance_m32}
\end{figure}

Finally, given this identification with the Mat\'ern 3/2 kernel, the most rigorous test of the performance is to compare the simulation on a new forcing signal as was done with the Mat\'ern 1/2. %
This is shown in \cref{fig:duff_sim_performance_m32} and a very similar pattern is observed. %
The nonlinear model vastly outperforms the linear model and its error is very low $<1\%$ NMSE. %
Again the observed error of $0.6\%$ NMSE is slightly over that seen in the case with the Mat\'ern 1/2 kernel but it is comparable and not sufficiently different to suggest that the kernel is playing a significant role. %

In this short appendix it has been discussed how alternative kernels may be used in the model. %
This was shown to have minimal effect on the performance of the model and the results were very closely aligned with those seen in the main body where a Mat\'ern 1/2 kernel is used. %
This outcome follows from what would be expected given the use of a GP, that the GP imposes only a prior on the function being identified and that the inclusion of the data through filtering and smoothing diminishes the importance fo the choice of kernel. %
This lack of sensitivity is reassuring to those wishing to apply this technique as it suggests that a full exploration of all possible kernels may not be required. %
It has been the experience of the authors that the Mat\'ern family, either 1/2 or 3/2, provide a good choice for a general purpose kernel in the application of LFM models.

\newpage
\section{Silverbox Prior Sensitivity} \label{sec:silverbox_priors}
\setcounter{figure}{0}  
\setcounter{table}{0}  

To explore the sensitivity of the solution for the silverbox system (\cref{sec:silverbox}) to the choice of prior, a number of different priors were run. %
In the main body, the prior distributions over the parameters are set using best practice, some estimation based on available engineering knowledge of the system. %
It is unknown if the priors are in fact a good representation of the system as the \emph{ground truth} is unknown and the priors chosen could be biased. %
In this sense, they represent a realistic scenario for a Bayesian model in engineering, where partial information is available but this information may be biased or incorrect.
Given this situation, there is a possibility that the estimated priors may be biased, it is worth exploring the effect this may have on the inference carried out. %
Deliberately, the priors were chosen to reflect a relatively large degree of uncertainty in the physical parameters of the model, see the diffuse nature of the priors shown in \cref{fig:silverbox_posteriors} relative to the posteriors. %
To test the sensitivity of the model to this prior selection the inference task was repeated with the means of those priors randomly perturbed, multiplying the mean value of each prior by the exponential of an independent standard Gaussian random variable. %
\Cref{tab:silverbox_priors} shows the values used for each prior, note that the variance (being large) is left constant throughout.
For comparison, the first of these five priors is set to be the same as the priors used in the main body of this work.

\begin{table}[h]
    \caption{Specification of different priors, all $\mathcal{N}(\cdot,\cdot)$\label{tab:silverbox_priors}}
    \centering
    \begin{tabular}{clcccccc}\toprule
    \multicolumn{2}{c}{Parameter} & \multicolumn{5}{c}{Prior Mean} & Prior Variance \\\midrule 
      & & 1 & 2 & 3 & 4 & 5 & All \\\midrule 
     $m$ &  $(\times 10^{-6})$      & 5.373 & 2.521 & 3.258 & 3.901 & 5.611 & 10 \\
     $c$ & $(\times 10^{-4})$      & 2.190 & 9.719 & 6.571 & 5.630 & 2.477 & 1\\
     $k$ & $(\times 10^{0})$                       & 0.993 & 1.117 & 0.160 & 4.407 & 9.032 & 0.5\\
     $\sigma$ & $(\times 10^{-3})$ & 5.000 & 4.467 & 7.034 & 2.513 & 10.5220 & 50\\
     $\ell$ & $(\times 10^{0})$       & 0.400 & 0.435 & 0.427 & 0.105 & 0.787 & 50\\
     $R$ &  $(\times 10^{-6})$ & 1.251 & 1.134 & 1.303 & 2.856 &0.762  & 5\\
    \bottomrule
    \end{tabular}
\end{table}

\begin{figure}[h]
    \centering
    \includegraphics[width=0.8\textwidth]{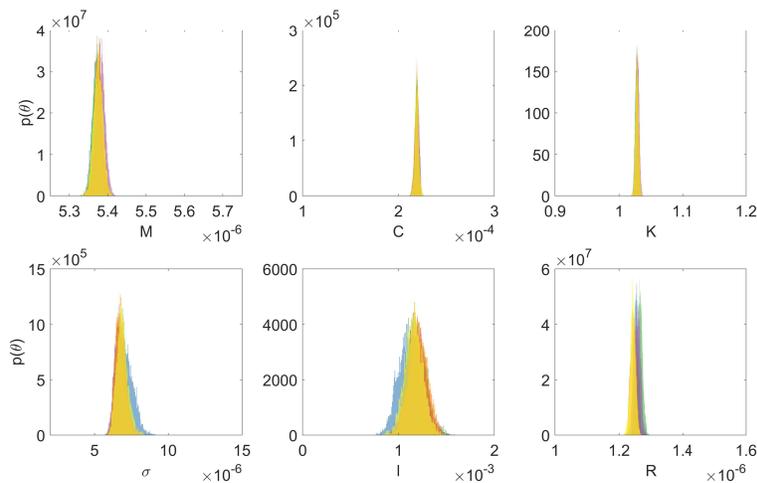}
    \caption{Parameter posteriors from the MCMC inference over the state-space model for the Silverbox under 5 different priors.}
    \label{fig:silverbox_posteriors_compare}
\end{figure}

Proceeding with inference as in the main-body text \cref{sec:silverbox}, the posteriors over the estimated parameters are shown in \cref{fig:silverbox_posteriors_compare}, with each prior shown using a different colour. %
In this figure, the posteriors over the parameters are shown for five different randomly perturbed priors. %
It can be seen that the posteriors have very little variation owing to the change in the priors. %
The priors are not shown in these figures (since they are not clearly visible owing to their high variance compared to the posterior), for details of the priors see \cref{tab:silverbox_priors}.
All of the priors have significantly larger variance that the posteriors showing that good information is recovered from the observed data. %
It is expected that not all of the posterior distributions will be identical, but it is reassuring that in the examples shown here the variation between them is not significantly affected by deliberately introducing large bias into the priors.

\end{document}